\documentclass{article}

\usepackage{pdfpages}
\usepackage{tablefootnote}
\usepackage{arxiv}
\usepackage[utf8]{inputenc} 
\usepackage[T1]{fontenc}    

\usepackage{url}            
\usepackage{booktabs}       
\usepackage{amsfonts}       
\usepackage{nicefrac}       
\usepackage{microtype}      
\usepackage{lipsum}
\usepackage{graphicx}
\graphicspath{ {images/} }
\usepackage{enumitem}
\usepackage{caption}
\usepackage{subcaption}
\usepackage{float}
\usepackage{mathtools}
\usepackage{amsmath}
\usepackage{algorithm}
\usepackage{algpseudocode}
\usepackage{array}
\usepackage{tikz}
\usepackage{tikz-dependency}
\usetikzlibrary{fit,positioning,arrows.meta,shapes}
\tikzset{neuron/.style={shape=circle, minimum size=1.25cm, 
  inner sep=0, draw, font=\small}, io/.style={neuron, fill=gray!20}}
\usetikzlibrary{shapes,arrows}
\usetikzlibrary{calc}
\usetikzlibrary{positioning}
\usepackage[symbol]{footmisc}

\title {Applications of Reinforcement Learning in Finance \\[1ex] \large Trading with a Double Deep Q-Network}

\author{
  Frensi Zejnullahu\\
  \small{School of Engineering}\\
  \small{Zurich University of Applied Sciences}\\
  \small{Winterthur, Switzerland}\\
  \small{\texttt{zejnufre@students.zhaw.ch}}
  \And
  Maurice Moser\\
  \small{School of Engineering}\\
  \small{Zurich University of Applied Sciences}\\
  \small{Winterthur, Switzerland}\\
  \small{\texttt{mosermau@students.zhaw.ch}}
  \And
  Joerg Osterrieder\footnotemark[1]\\
  \small{School of Engineering}\\
  \small{Zurich University of Applied Sciences}\\
  \small{Winterthur, Switzerland}\\
  \small{\texttt{joerg.osterrieder@zhaw.ch}}
   }

\begin{document}
\maketitle

\begin{abstract}

This paper presents a Double Deep Q-Network algorithm for trading single assets, namely the E-mini S\&P 500 continuous futures contract. We use a proven setup as the foundation for our environment with multiple extensions. The features of our trading agent are constantly being expanded to include additional assets such as commodities, resulting in four models. We also respond to environmental conditions, including costs and crises. Our trading agent is first trained for a specific time period and tested on new data and compared with the long-and-hold strategy as a benchmark (market). We analyze the differences between the various models and the in-sample/out-of-sample performance with respect to the environment. The experimental results show that the trading agent follows an appropriate behavior. It can adjust its policy to different circumstances, such as more extensive use of the neutral position when trading costs are present. Furthermore, the net asset value exceeded that of the benchmark, and the agent outperformed the market in the test set. We provide initial insights into the behavior of an agent in a financial domain using a DDQN algorithm. The results of this study can be used for further development.

\end{abstract}

\textbf{Keywords:} Deep Reinforcement Learning $\cdot$ Double Deep Q-Network $\cdot$ Trading Agent

\footnotetext[1]{
Financial support by the Swiss National Science Foundation within the project “Mathematics and Fintech - the next revolution in the digital transformation of the Finance industry” is gratefully acknowledged by the corresponding author. 
This research has also received funding from the European Union's Horizon 2020 research and innovation program FIN-TECH: A Financial supervision and Technology compliance training program under the grant agreement No 825215 (Topic: ICT-35-2018, Type of action: CSA) and from Innosuisse under the grant agreement "Strengthening Swiss Financial SMEs through Applicable Reinforcement Learning" (Innosuisse Innovation project 47959.1 IP-SBM).
Furthermore, this article is based upon work from the COST Action 19130 Fintech and Artificial Intelligence in Finance, supported by COST (European Cooperation in Science and Technology), www.cost.eu (Action Chair: Joerg Osterrieder).\newline
The authors are grateful to the management committee members of the COST (Cooperation in Science and Technology), Action Fintech, and Artificial Intelligence in Finance as well as speakers and participants of the $4^{\text{th}}$, $5^{\text{th}}$, and $6^{\text{th}}$ European COST Conference on Artificial Intelligence in Finance and Industry.}

\newpage
\tableofcontents
\newpage

\section{Introduction}
Reinforcement learning (RL) is besides Supervised and Unsupervised Learning the third branch of the machine learning paradigm. Compared to Supervised and Unsupervised Learning, RL takes a different approach where the core functionality involves the interaction between an agent and its environment to maximize a numerical reward signal \cite{sutton2018reinforcement}. Recent advances such as deep learning have had a significant impact on Reinforcement Learning applications. It dramatically improved the state-of-the-art in solving tasks and showed superhuman trading performances \cite{2202.04115}. A key focus of RL in finance is the development of fully autonomous systems that mimic a financial advisor in terms of analysis and decision making for financial tasks such as trading.

Over the past decade, advances in Deep Learning have been made thanks to big data and increasing computing power \cite{1810.06339}. Deep Reinforcement Learning (DRL) in finance \cite{2011.09607} can be used by agents to find their optimal policy in order to receive the maximal cumulative reward or return. The agent gets information about the current state and tries to approximate its value using a neural network, which should lead to an optimal decision or action \cite{2112.04755}. The only way to find the optimal policy is to evaluate the actions taken by the agent. There are some common challenges when approximating a non-linear function for an improved policy. Some of these challenges for automated financial trading are:

\begin{enumerate}
    \item \textbf{Credit Assignment:} 
    In reinforcement learning, delayed feedback is widespread. The trading return for an agent may occur much later than the associated action. Therefore, the question arises as to how the RL agent can recognize that specific trading strategies perform better than others if it only learns from a later feedback. This is referred to as the credit assignment problem \cite{jansen2020machine}.
    
    \item \textbf{Exploration vs. Exploitation:} 
    The financial market as an environment in RL is only partially observable by an agent \cite{2202.11296}. Consequently, trading decisions are based on an incomplete learning process in which the agent has to find the optimal strategy. This poses the risk that it exploits previous successful trading strategies instead of exploring new strategies that may be more suitable. Therefore, the RL agent must balance this tradeoff between exploration and exploitation \cite{jansen2020machine}.
    
    \item \textbf{Financial Aspect:} 
    Many market fluctuations add a lot of noise to the financial data \cite{Ye_Pei_Wang_Chen_Zhu_Xiao_Li_2020}. Thus, training on this data can be challenging for the RL agent. Furthermore, RL applications in finance provide little guidance on the hyperparameters used in the models.
\end{enumerate}

All of these challenges aim to achieve an optimal estimate based on the value of a state. There are policy-based and value-based methods \cite{2005.14419} that can be applied to a model-free setting such as the financial market.\cite{2112.04553}. On the one hand, for instance, the Policy Gradient Method (policy-based), approximates the policy directly (action probabilities). On the other side (value-based), a Q-Learning method can be used to approximate an action-value function or better known as a Q-function \cite{NIPS1999_464d828b}. Q-Learning was a breakthrough for RL in 1989 \cite{NIPS1999_464d828b} and is currently popular in combination with Deep Learning in finance. The Deep Q-Network (DQN) is a model-free algorithm that estimates the Q-function using neural networks. The basic DQN architecture has been improved for efficiency and higher performance. Currently, the Double Deep Q-Network (DDQN) is state of the art, showing better estimation compared to the general DQN approach.

Concerning Stefan Jansen's work \cite{jansen2020machinelearning}, we apply an adapted DDQN algorithm to single asset trading; the S\&P 500. We use the transformed 1-day and 5-day returns of the S\&P 500 as input for the RL agent and additionally the trading strategy "Long and Hold" as the benchmark for our agent performance. In the next step, we include additional features from other assets. We stepwise add the features (1-day and 5-day returns) of the Russel 2000, Oil, and Gold assets to the input for the RL model. We use the asset returns over the last 15 years, split into training and test data. {\bf The primary purpose of this paper is to compare the performance of the RL agent in terms of adding additional features.} To satisfy the comparison, we keep the hyperparameters constant for all models.

The remainder of the paper is structured as follows: Section 2 briefly presents the current and used literature. Our methodology, proposing the theory of an optimal policy in the context of the DDQN algorithm, is shown in Section 3. Section 4 explains the experiment with the data, the RL settings and the following results. Section 6 discusses the conclusions and the implications for an outlook.

\clearpage
\section{Literature Review}
The following studies on reinforcement learning are related to our work and can be divided into three parts. The first part covers the theoretical explanation of the different RL properties and approaches we use to point to the concept of the DDQN algorithm. In the second part, we present the achievements of Deep Reinforcement Learning in general \cite{2201.02135}, followed by several papers focusing on the performance of an RL agent in a financial setting, combined with neural networks. The third and last part can be considered the basis for our experiment's realization.

The Introduction to Reinforcement Learning by Sutton and Barto \cite{sutton2018reinforcement} covers almost all concepts of RL and demonstrates them with examples. It can be treated as a reference book or as a theory wrapped in lectures. It explains RL in a discrete or stochastic environment, model-free or model-based, and the three well-known solution methods, Dynamic Programming, Monte Carlo Methods and Temporal Difference Learning. We mainly used the theoretical aspects of Temporal Difference Learning and their optimal policy concerning the Q-value function. Function approximators for optimal policies are essential in RL, and Sutton et al. \cite{NIPS1999_464d828b} reveal a different approach. They illustrate actor-critic methods based on the Policy-Gradient Theorem and their approximation. In addition, Hambly et al. \cite{2112.04553} provide an overview of recent advances in RL, focusing on deep reinforcement learning and applications in finance. This involves using a large amount of financial data with fewer model assumptions, which can improve decisions in complex financial environments. Furthermore, the value and policy based methods can be adapted for neural network approximation. We consider this work as a bridge for applying neural networks in a RL framework (Deep Reinforcement Learning). 

The Deep Q-Learning algorithm (DQN) can learn successful policies directly from high-dimensional sensory inputs. Mnih et al. \cite{Mnih2015HumanlevelCT} tested the agent on 2600 classic Atari games and received only the pixels and the game scores as inputs. The DQN algorithm outperformed all previously used algorithms and reached the professional human level in 49 Atari games. This results in the first artificial agent capable of handling many challenging tasks. Continuing with these results, Hasselt et al. \cite{1509.06461} introduced the Double Deep Q-Network (DDQN). They show that the DQN algorithm suffers from significant overestimation for some games in the Atari 2600 domain. Extending the DQN to the DDQN reduces overestimation and results in substantially better performance across multiple games.

A DQN application for building long-short trading strategies in finance is presented by Hirsa et al. \cite{2106.08437}. Training is performed using artificial data and actual price series to trade the E-mini S\&P 500 futures contract. The authors conclude that the generation of artificial financial data is of great importance and needs to be supported by insights into the time series behavior of financial markets. Portfolio management and trading involves the analysis of financial assets and the estimation of expected future returns and risks. The DQN algorithm by Pigorsch and Schäfer \cite{2112.04755} is able to trade high-dimensional portfolios from cross-sectional datasets. The out-of-sample analysis of 48 portfolio setups (only US stocks) far outperforms both passive and active benchmark investment strategies on average. Furthermore, Huang \cite{1807.02787} provides an additional feedback signal for all actions for the trading agent, allowing for a greedy policy instead of the standard $\varepsilon$-greedy policy. For all currency pairs, the agent outperforms the two benchmark strategies (staying long/short). The author concludes that higher trading costs do not actually reduce the performance as trained strategies become more robust. Overall, these studies demonstrate the potential of Deep Reinforcement Learning as a future tool for financial analysis.

The reference model used in our experiment is based on the work of Stefan Jansen \cite{jansen2020machinelearning}. In chapter 22, the author moves from traditional Supervised and Unsupervised approaches to Reinforcement Learning \cite{jansen2020machine}. The chapter opens with key concepts and examples of conventional RL, followed by implementing neural networks using the open-source package Open AI Gym \cite{brockman2016openai}. In the second implementation, an agent trades a single asset in the stock market using a DDQN architecture. The comparison with the buy-and-hold benchmark shows that the agent can achieve better returns after a certain number of training episodes. Hambly et al. \cite{2112.04553} presents the use of Reinforcement Learning in various aspects of finance. The author provides initial insights, followed by familiar concepts such as RL in portfolio optimization, and concludes the usage with Smart Order Routing. 

Our research belongs to the financial topic of "Optimal Execution" in chapter 4.3 of Hambly et al. \cite{2112.04553}. In the existing literature on RL for optimal execution, a number of algorithms have been used. The most popular types were various Q-learning algorithms (DQN \cite{1911.10107}, DDQN \cite{1812.06600}) and policy-based algorithms \cite{2011.10300} such as A2C, PPO, and DDPG \cite{ye2020optimal}. We contribute to the existing literature on RL applications in finance by presenting a DDQN algorithm for trading S\&P 500 with adjustable number of features. The base model \cite{jansen2020machinelearning} is extended in the following main directions: First, data preparation becomes more financially oriented in terms of normalization, time series length, and hyperparameter. Second, the features are extended to multiple assets and multiple models to be trained. Finally, we introduce a test set to measure the out-of-sample performance. We also refrain from adjusting the models for a second training process so that the agents' trading strategies are kept unbiased.

\section{Methodology}
This section covers the theoretical background of the application of the DDQN algorithm. Some key concepts leading to optimal policy are introduced, followed by Q-Learning, the underlying framework of our DDQN. It closes with a comparison of the DQN algorithm and its extension to the DDQN.

\subsection{Temporal Difference Learning} \label{tdl}
There are several approaches to solve a Reinforcement Learning problem. All of them involve finding an optimal behavior for the agent. The three most common approaches are Dynamic Programming (DP) \cite{2109.11808}, Monte Carlo (MC) \cite{2002.03585}, and Temporal Difference (TD) \cite{jansen2020machine}. TD Learning can be viewed as a combination of the MC and DP approaches. TD and MC methods do not need to know the dynamics of the model and therefore use only experience to solve prediction problems. On the other hand, TD and DP methods update their estimates based on other estimates during the process (sequences) and do not have to wait until the final result. TD relies on bootstrapping whereas MC samples the whole sequence to estimate a value \cite{sutton2018reinforcement}. The Q-Learning method we use for our trading agent is based on an off-policy TD algorithm. 

\subsection{The Agent's Policy}\label{agent_policy}
The following paragraphs outline the basis for an optimized policy.

\paragraph{Return:}The behavior of our trading agent is associated with the total return it achieves \cite{1905.11591}. In practice, the trading agent observes some information in the environment called state $S_t$, chooses an action $A_t$ based on its policy, and finally receives a reward (return) $R_{t+1}$. The interaction between our agent and the environment generates a trajectory $\tau$ = [$S_t$, $A_t$, $R_{t+1}$, $S_{t+1}$, $A_{t+1}$, $R_{t+2}$, $\dots$]. In general, the goal of our trading agents is to behave in a way to maximize the cumulative returns. The sequence of rewards $R_{t+1}$, $R_{t+2}$, $R_{t+3}$, $\dots$ that needs to be maximized can be defined as the expected future return $R_{t}$, denoted as $G_t$:
\begin{equation} \tag{3.1}\label{eq:3.1}
    G_t = E[R_{t+1} + \gamma R_{t+2} + \gamma^2 R_{t+3} + \dots] 
= \sum_{s=0}^{T} \gamma^{s} E[R_{t+s}]
\end{equation}

where $ 0  \le \gamma \le 1 $ is the discount factor \cite{2007.02040} to include the time value. In addition, the return function can be expressed recursively:
$$
G_t = R_{t+1} + \gamma R_{t+2} + \gamma^2 R_{t+3} + \dots \newline
$$
$$
G_t = R_{t+1} + \gamma (R_{t+2} + \gamma R_{t+3} + \dots) \newline
$$
\begin{equation} \tag{3.2}\label{eq:3.2}
G_t = R_{t+1} + \gamma G_{t+1}.
\end{equation}

This type of relationship is used repeatedly in the formulation of RL algorithms and is of great importance \cite{jansen2020machine}.

\paragraph{Value Function:}Most RL models include an estimation of a value function. The value function is crucial for the agent as it should lead to an optimal policy \cite{2005.10804}. It estimates the expected future return for every state or state-action pair. In our case, we are estimating the value function for every state-action pair that can be seen as performing a particular action being in a given state. We are talking about a Markov Decision Process (MDP) \cite{2201.05000} where the policy $\pi$ of the agent maps all states to probability distribution given each possible action. Therefore, the probability of performing an action $a$ in a state $s$ at time t can be viewed as $\pi(a|s) = P(A_t = a | S_t = s)$. We follow the definition of the state-action value function, denoted $q(s,a)$, as the expected total return starting in a given state $s$, taking action $a$ and following the policy $\pi$ \cite{sutton2018reinforcement}:
\begin{equation} \tag{3.3}\label{eq:3.3}
q_\pi(s,a) = E_\pi[G_t | S_t = s, A_t = a] = E_\pi[\sum_{k=0}^{\infty} \gamma^{k} R_{t+k+1} | S_t = s, A_t = a].
\end{equation}

\paragraph{The Bellman Equations:} The value function is characterized by a key property: It can be defined recursively such as the return function in \ref{eq:3.2}. The Bellman Equations expresses the recursive relationship between the value of the current state and the value of its successor state. This is indicated in the following equation for our state-action value function\cite{sutton2018reinforcement}:
$$
q_\pi(s,a) = E_\pi[G_t | S_t = s, A_t = a] = E_\pi[R_{t+1} + \gamma q_\pi(s',a')]
$$
\begin{equation} \tag{3.4}\label{eq:3.4}
q_\pi(s,a) = \sum_{s',r}p(s',r | s,a)[r + \gamma \sum_{a'}\pi(a'|s')q_\pi(s',a')].
\end{equation}
The {\bf Bellman Expectation Equation} \ref{eq:3.4} states that under policy $\pi$, the value of a state $s$ is equal to the value of its successor state plus the reward earned in that new state \cite{jansen2020machine}.

\paragraph{Optimal Policy:} To maximize the cumulative reward in the long run, the agent needs to optimize its policy $\pi$. The derivation of an ideal policy is closely related to the optimization of our state-action value function \cite{2104.08805}. An optimal policy, denoted $\pi_*$ is defined to be better than or equal to a policy $\pi'$ if its expected return for all states is at least as high as that of policy $\pi'$. To put it in our terms , $\pi_* \ge \pi'$ if and only if $q_*(s,a) \ge q_{\pi'}(s,a)$ for all $s \in S$. The optimal policy that can be viewed as the optimal state-action value function, denoted $q_*(s,a)$ is defined as \cite{sutton2018reinforcement}:

\begin{equation} \tag{3.5}\label{eq:3.5}
q_*(s,a) = \mathop{max}_{\pi}q_\pi(s,a).
\end{equation}

This optimal state-action value function can also be expressed recursively as in \ref{eq:3.4} by the Bellman equations as follows:
$$
q_*(s,a) = E[R_{t+1} + \gamma\mathop{max}_{a' \in A} q_*(S_{t+1},a') | S_t = s, A_t = a]
$$
\begin{equation} \tag{3.6}\label{eq:3.6}
q_*(s,a) = \sum_{s',r}p(s',r | s,a)[r + \gamma\mathop{max}_{a' \in A} q_*(s',a')].
\end{equation}
The {\bf Bellman Optimality Equation} \ref{eq:3.6} shows that the value of any state within an optimal policy must equal the expected total return given the best action. The general concept of many RL algorithms is to approximate the given value function using the Bellman equations as an iterative update. These iterative updates usually converge towards the optimal value function $q_*$ \cite{Mnih2015HumanlevelCT}.

\paragraph{Q-Learning:} In the model-free environment for our trading agent, the transition probabilities in the optimal state-action value function \ref{eq:3.6} are unknown. As mentioned in Section \ref{tdl}, Q-Learning is based on Temporal Difference Learning (TD), which stochastically approximates the Bellman optimality equation \ref{eq:3.6} without knowing these transition probabilities \cite{2112.04553}. Moreover, the Q-Learning algorithm follows an off-policy, implying that it does not strictly stick to the given policy. This is realized with an $\varepsilon$-greedy policy. Our agent performs a random action with probability $\varepsilon$ or the best action as defined by the value function with probability $1-\varepsilon$ \cite{jansen2020machine}. The Q-Learning algorithm is defined as:

\begin{equation} \tag{3.7}\label{eq:3.7}
Q(S,A)_{new} = (1-\alpha) \underbrace{Q(S,A)}_\text{current estimate} + \ \alpha \ [\underbrace{R + \gamma\mathop{max}_{a' \in A}Q(S',a')}_\text{new estimate (TD target)}]
\end{equation}

where $ 0  \le \alpha \le 1 $ is the learning rate. The new estimate consists of the return on the $\varepsilon$-greedy policy and the discounted value function for the next step. The Q-Learning algorithm updates its one-step look-ahead estimate using bootstrapping based on experience \cite{jansen2020machine}.

\subsection{Deep Reinforcement Learning}\label{drl}

The Q-Learning algorithm is a valuable tool for policy optimization in a discrete space environment. Since our environment is data from the financial market, the state space is continuous while the action space remains discrete. A continuous state space estimates the optimal policy with traditional approaches more complex and often not efficient. Therefore, we combine Q-Learning with a deep neural network (DNN), and instead of using Q-tables, we use the DNN to approximate our optimal value function $Q(S,A; \theta)$ with $\theta$ as the weights for our DNN \cite{2112.04553}.

\paragraph{Deep Q-Network:} 
The Deep Q-Network (DQN) \cite{1312.5602} uses a deep convolutional neural network with the state space as an input and the action space as the output. The DQN is trained to adjust the weights $\theta_i$ at every iteration $i$ in order to reduce the mean square error in the Bellman equation \cite{Mnih2015HumanlevelCT}. A Loss function based on stochastic gradient descent (SGD) is used to calculate the squared difference between the target estimation and the current Q-value estimation $Q(s,a;\theta)$ \cite{jansen2020machine}. The target estimation relies on \ref{eq:3.7} and is described as:
\begin{equation} \tag{3.8}\label{eq:3.8}
y_i = r + \gamma\mathop{max}_{a' \in A} Q(s',a';\theta_i| s, a)
\end{equation}
and following the notation of the Loss function:
\begin{equation} \tag{3.9}\label{eq:3.9}
L_i(\theta_i) = (\ \underbrace{E[y_i]}_\text{Q-target} - \underbrace{Q(s,a;\theta_i)}_\text{current prediction})^2.
\end{equation}
Typically, applying a nonlinear function approximator such as a neural network in an RL environment is very unstable. This is due to correlations in the data sequence or to the fact that minor updates of the Q-network can substantially change the policy \cite{Mnih2015HumanlevelCT}. To overcome these instabilities, the DQN architecture has been improved in several ways. Two of them are essential to mention:

\begin{enumerate}
    \item \textbf{Experience Replay \cite{2007.06700}:} In Experience Replay, the DQN algorithm is extended by a replay memory $M$. On the one side, the replay memory stores tuples $(s_t, a_t, r_t, s_{t+1})$ at each time step. On the other side, mini-batches of these tuples are randomly selected from $M$ to update the neural network weights before an action is executed. Since time series sequence is usually correlated, experience replay aims to reduce this correlation by inserting mini-batches of tuples. The size of $M$ is generally very large, and a randomly selected sample is consequently nearly independent \cite{2112.04553}.
    
    \item \textbf{The Target Network:} Instead of applying the DQN with only the online network $Q(s,a;\theta)$, we add a second network, the target network $Q(s,a;\theta')$ with weights $\theta'$ to the algorithm. The target network can be considered as a copy of the online network, with the difference that the weights are updated only after $\tau$-steps. We now use the target network with its weights $\theta'$ to estimate the Q-target $y_i$ \cite{1701.07274} in \ref{eq:3.8} and \ref{eq:3.9}:
    \begin{equation} \tag{3.10}\label{eq:3.10}
    y_i = r + \gamma\mathop{max}_{a' \in A} Q(s',a';\theta'_{i}| s, a).
    \end{equation}
    The fact that the target network is fixed for $\tau$-steps and only then updated with the current weights $\theta$ of the online network prevents instabilities from spreading quickly. Moreover, it reduces the risk of divergence when approximating an optimal policy \cite{2112.04553}.
\end{enumerate}
These two critical improvements to the DQN algorithm have significantly increased performance without making it noticeably slower \cite{1509.06461}.

\paragraph{Double Deep Q-Network:} 
The DQN algorithm estimates the Q-target using a max operator \ref{eq:3.10}, intentionally taking the highest value. It not only performs the action, but also evaluates it based on this procedure. The selection and evaluation of the highest value has been shown to be overestimated, which can lead to stagnation of the training process. The Double Deep Q-Network (DDQN) \cite{1509.06461} decouples the selection and evaluation of the action to prevent this bias. In this process, the action is performed based on a network with weights $\theta$, while the action is evaluated with a second network with weights $\theta'$ considering the next state, which can be formally denoted as follows \cite{NIPS2010_091d584f}:
\begin{equation} \tag{3.11}\label{eq:3.11}
y_i = r + \gamma Q(s',\mathop{argmax}_{a' \in A}Q(s', a, \theta_{i});\theta'_{i}| s, a).
\end{equation}
The DQN algorithm already uses a second network (target network) with weights $\theta'$, which can be viewed as a natural choice for the DDQN algorithm. In conclusion, the DDQN is an extension of the DQN, with the key feature that it additionally uses the target network to separate the execution and evaluation process of an action.

\section{Experiment}
In this section, we first present the financial market data, followed by the models and the features of our trading agent. This includes the specification of our algorithm and hyperparameters. We finish the chapter with the results of the trading process.

\subsection{Data}
The data used in the models include the S\&P 500, Russel 2000, WTI Crude Oil, and Gold commodity, either as continuous futures contracts or ETF, all in USD. The data is retrieved from the Refinitiv database, which covers 2007 to 2022 \cite{refinitiv}. 

\begin{table}
\renewcommand{\arraystretch}{1.5}
\centering
\caption{Asset retrieval from Refinitiv}
\begin{tabular}{ l | l | l | l }
\hline
Class           & Exchange      & Symbol    & Name    \\\hline
Equity Index    & CME           & ESc1      & S\&P 500 E-Mini Future     \\
                & NYSE Arca     & IWM       & Russel 2000 iShares ETF   \\\hline
Commodity       & COMEX         & GCc1      & Gold Future               \\
                & NYMEX         & CLc1      & WTI Crude Oil Future      \\\hline
\end{tabular}
\label{tab:data}
\end{table}

The data preparation includes the conversion of the prices into normalized returns \cite{2003.00598}, dropping any row with missing values, and the separation into a train and test set. As input data, the log returns are scaled by their respective standard deviation. The normalized daily returns are termed as $r_{t}^{(i)} /\left(\sigma_{t}^{(i)} \sqrt{252}\right)$ with the exponential weighted moving average of the standard deviation $\sigma_{t}^{(i)}$. Therefore, the normalized data serve as states for the trading agent, corresponding to features for a neural network.

\subsection{Model} \label{model}
To compare the performance of the additional features for the RL agent, we will present four models. They each take the normalized 1-day and 5-day returns of the S\&P 500 as the underlying.

\begin{itemize}
    \itemsep0em
    \item Model 0: 1-day/5-day return {\bf S\&P 500}
    \item Model 1: 1-day/5-day return {\bf S\&P 500}, {\bf Russel 2000}
    \item Model 2: 1-day/5-day return {\bf S\&P 500}, {\bf Russel 2000}, {\bf WTI}
    \item Model 3: 1-day/5-day return {\bf S\&P 500}, {\bf Russel 2000}, {\bf WTI}, {\bf Gold}
\end{itemize}

\subsection{Environment}

We now proceed to the definition of the three main attributes of the trading agent: the state space, the action space, and the reward function.

\paragraph{State Space:}
As previously mentioned, the trading agent is given a sample of the environment, which can be viewed as its state. We are talking about a stochastic and continuous state space. Thus, the state is the normalized 1-day and 5-day asset returns. In the simplest model, only the return on the S\&P 500 itself is included (2 inputs), while in the other models, the other assets will be added. For example, Model 3 features the 1-day and 5-day returns of all assets (8 inputs).
\paragraph{Action Space:}
Based on the given state at time $t$, the trading agent will predict the direction of the financial market at $t+1$. We therefore use a discrete action space in which three positions are represented as:
\begin{equation} \tag{4.1}\label{eq:4.1}
  A_t =
    \begin{cases}
      \quad0, \, & \text{short signal}\\
      \quad1, \, & \text{stay-out-of-the-market signal}\\
      \quad2, \, & \text{buy signal}
    \end{cases}       
\end{equation}
where action 0 (stay-out-of-the-market) implies that any open position will be closed. In addition, there are trading and time costs coupled with number of actions. For example, changing the position from -1 to 1 (and vice versa) corresponds to two trades (2 times the trading cost), while no change in position is associated with time costs.

\paragraph{Reward Function:}
The third part in the RL framework is the reward function \cite{2205.15400}, from which our agent learns and tries to maximize its output. We define it as the product of our action and the return of the S\&P 500 deducted by the trading costs:
    \begin{equation} \tag{4.2}\label{eq:4.2}
    R_t = R_{t, market}*(A_{t-1} - 1) - \text{(trading costs $+$ time costs).}
    \end{equation}
The actions are decremented by 1, that translates to \{-1, 0, 1\} for a proper multiplication of the positions with the return. To compare the performance of the trading agent with a benchmark, we use the net asset value (NAV), which is defined as the cumulative reward per day.

\subsection{DDQN Algorithm}
The extension of the DQN algorithm to the DDQN algorithm allows the agent to develop an optimal behavior with better performance \cite{1509.06461}. In short, the agent transmits the 1- and 5-day returns as a state to the DDQN, which estimates and returns the current best action \{0, 1, 2\} based on its policy. The action is then performed and evaluated with the earned reward by the DDQN. The DDQN with its properties is therefore considered the core of the RL framework, and the following algorithm describes the process adapted to our case:

\begin{algorithm}[H] 
\caption{Double Deep Q-Learning for trading a single asset}
Extensions of the DDQN algorithm from Hasselt \cite{NIPS2010_091d584f} are highlighted in italics.\\
\label{alg:cap}
\begin{algorithmic}[1]
\State Initialize replay memory $\mathcal{D}$ to capacity $\mathcal{N}$
\State Initialize online network $Q_{\theta}$, target network $Q_{\theta'}$

\For{episode = 1 to $M$}
    \State \emph{Reset trading environment}
    \State \emph{Initialize Net Asset Value (NAV) = 0}
    \State Initialize $s_m = \{r_{1}, r_{5}\}$
    \For{t = 1 to T}
        
        \State Observe state $s_t = \{r_{1}, r_{5}\}$
        \State With probability $\varepsilon$ select random action $a_t$ otherwise select $a_t \sim \pi(a_t,s_t)$
        \State Execute action $a_t$ and observe next state $s_{t+1}$ 
        \State \emph{Update trading position and receive reward $r_t$ according to (4.2)}
        \State \emph{Add reward to NAV}
        \State Store transition $(s_t, a_t, r_t,s_{t+1})$ in replay memory  $\mathcal{D}$
        \State sample mini-batch $e_t = (s_t, a_t, r_t, s_{t+1}) \sim \mathcal{D}$
        \State Compute target Q value $Q^*(s_t,a_t)$ according to (3.1)
        \State Perform gradient descent step on $(Q^*(s_t,a_t) - Q_\theta(s_t,a_t))^2$
        \For{every $\tau$ step}
            \State Update target network parameters: $\theta' \leftarrow \theta$
            \EndFor
        \EndFor
    \EndFor
\end{algorithmic}
\end{algorithm}

\subsection{DDQN Architecture}
The DDQN architecture maps two features (4, 6, and 8 for Model 1-3) to three outputs (target) to represent the Q-value for each action, which can then be selected based on the policy. Our DDQN architecture includes 2 layers of 64 units, each with L2 activity regularization and a dropout layer. The number of trainable parameters is manageable with 4547 for the base model shown in Table \ref{tab:2}.
\begin{table}
\renewcommand{\arraystretch}{1.5}
\centering
\caption{Model architecture for DDQN Model 0}
\begin{tabular}{ |l|c|c| }
\hline
Layer           & Output        & Parameter \\\hline
Input Layer     & 2             & 0         \\
Hidden Layer 1  & 64            & 192       \\
Hidden Layer 2  & 64            & 4160      \\
Dropout Layer   & 64            & 0         \\
Output          & 3             & 195       \\\hline
                & {\bf Total}         & {\bf 4547}      \\\hline
\end{tabular}
\label{tab:2}
\end{table}
The architecture for the other models (Model 1-3) remains the same, except for a higher input that increases the number of trainable parameters up to 4931. This can still be considered minimal number of parameters. In addition, the DDQN incorporates two neural networks. The second network (target network) is, as mentioned in Section \ref{drl}, a delayed copy of the online network and therefore also uses the same architecture.

\subsection{Hyperparameters}
The use of RL in Atari games has shown to be very successful \cite{Mnih2015HumanlevelCT}. One drawback of RL in finance is that there is no proven baseline or reference of hyperparameters to solve the model. Therefore, finding the optimal hyperparameters in a very time consuming RL environment can be quite difficult. The settings we use are oriented on the procedure in Stefan Jansen's work \cite{jansen2020machinelearning} with minor changes. In order to compare the individual performances, we kept the same hyperparameters for all four models. The exploration probability $\varepsilon$ is defined by a function where $\varepsilon$ decreases linearly at first and exponentially at the end from 1 to 0.01. All key parameters are summarized in Table \ref{tab:3} and Table \ref{tab:4}, and the description can be found in the appendix section \ref{hyper}.

\begin{table}
    \begin{minipage}{.49\linewidth}
        \renewcommand{\arraystretch}{1.5}
        \caption{Hyperparameters for all models}
        \centering
        \begin{tabular}{| l | c | c |}
            \hline
            Parameter                   & Symbol            & Value         \\\hline
            Episodes                    & $M$               & 1000          \\
            Episode length              & $T$               & 252           \\
            Discount factor             & $\gamma$          & 0.9           \\
            Learning rate               & $\alpha$          & 1e-4          \\
            Exploration probability     & $\varepsilon$     & [1,...,0.01]  \\
            Update target network       & $\tau$            & 100           \\\hline
        \end{tabular}
    \label{tab:3}
    \end{minipage}
    \begin{minipage}{.49\linewidth}
        \renewcommand{\arraystretch}{1.5}
        \caption{Hyperparameters for all models}
        \centering
        \begin{tabular}{| l | c |}
            \hline
            Parameter                   & Value         \\\hline
            Replay capacity             & 1e6           \\
            Batch size                  & 4096          \\
            Optimizer                   & Adam          \\
            Trading cost                & 1e-4 or 0     \\
            Time cost                   & 1e-5 or 0     \\\hline
        \end{tabular}
    \label{tab:4}
    \end{minipage} 
\end{table}

\subsection{Training}
The open source trading environment \cite{brockman2016openai} as well as the DDQN algorithm are tailored to our needs and are used for training our models. The procedure for training the RL agent is the same for each model. A training episode includes 252 steps (days) that can be counted as one trading year. The start date for the training of our agent is randomly chosen in each episode and lies between 2007 and 2019. At the start of each episode, the NAV of the agent and the market is set to 0. After one trading year, we store the NAVs and other performance measures for comparison purposes. The outperformance of an agent is given when the final NAV after 252 days is higher than that of the market. The training usually consists of 1000 episodes, which is equivalent to exercising trades for 1000 years. To counteract overfitting, we set a threshold where the agent outperforms the market. Training is terminated prematurely if the agent outperforms the market for 25 consecutive years (episodes). The out-of-sample performance of our models begins after the end of the train series and continues for either two or three trading years. It will trade the specific time period only once to remain out-of-sample. All models, including market performance, will be compared.

\subsection{Results}

We trained and tested four models for each scenario and compared them to the market (long and hold) as a benchmark. The models differ in the number of features, and the specifications can be found in in Section \ref{model}. The settings involve the models with and without transaction costs and the inclusion of the 2020 crisis in either the training or test set. We demonstrate the results in a scenario where the crisis is included in the test set. Before comparing the different models, we show the difference in the performance and behavior of our trading agent in terms of costs. For the following figures \ref{fig:1}, \ref{fig:2}, we intentionally increased the trading and time fees for our agent in Model 1 to 10 bps and 1 bps, respectively, to demonstrate behavior in a high cost environment.

\begin{figure}[ht]
     \centering
     \begin{subfigure}[b]{0.49\textwidth}
         \centering
         \includegraphics[width=\textwidth]{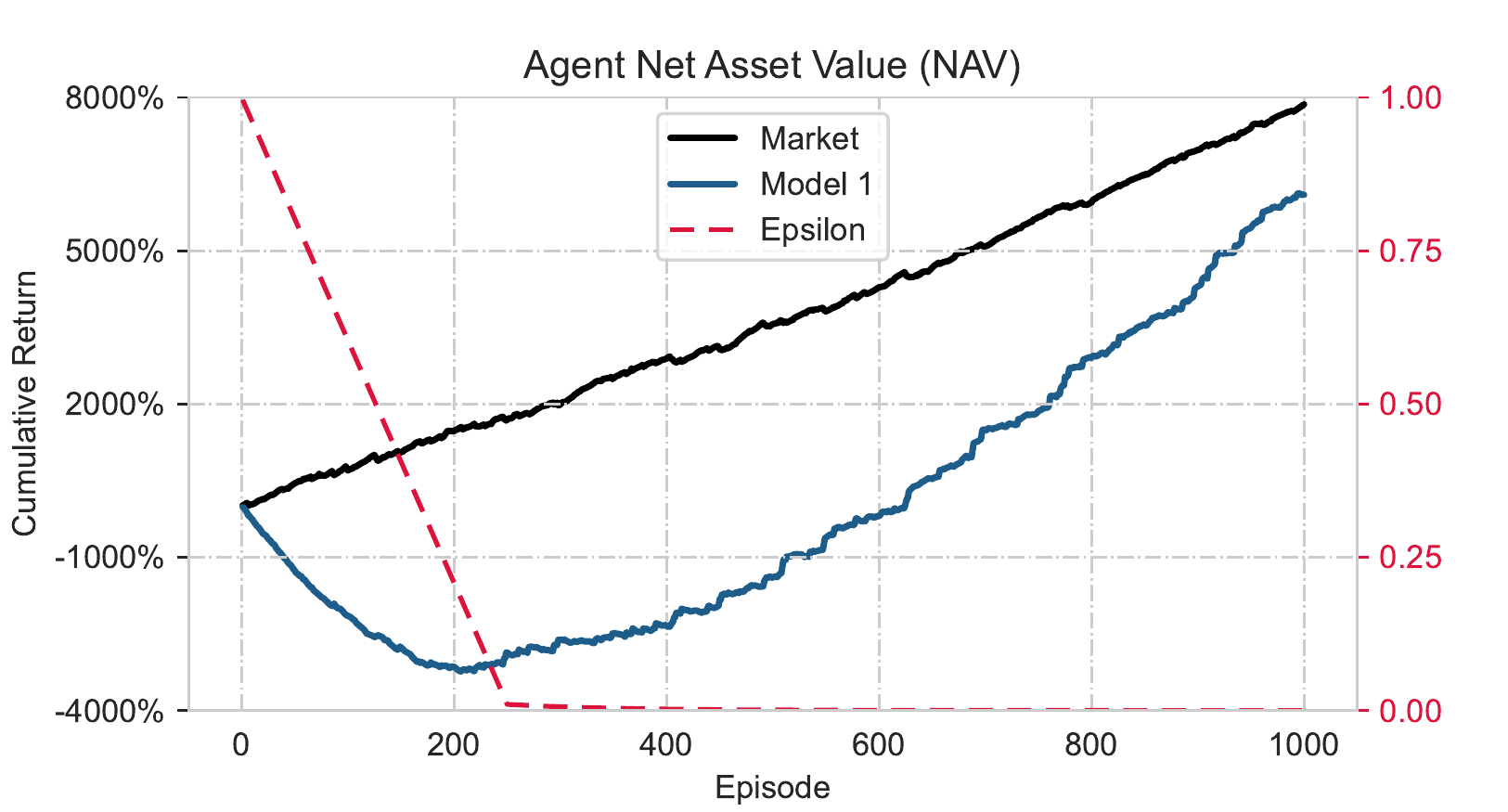}
     \end{subfigure}
     \begin{subfigure}[b]{0.49\textwidth}
         \centering
         \includegraphics[width=\textwidth]{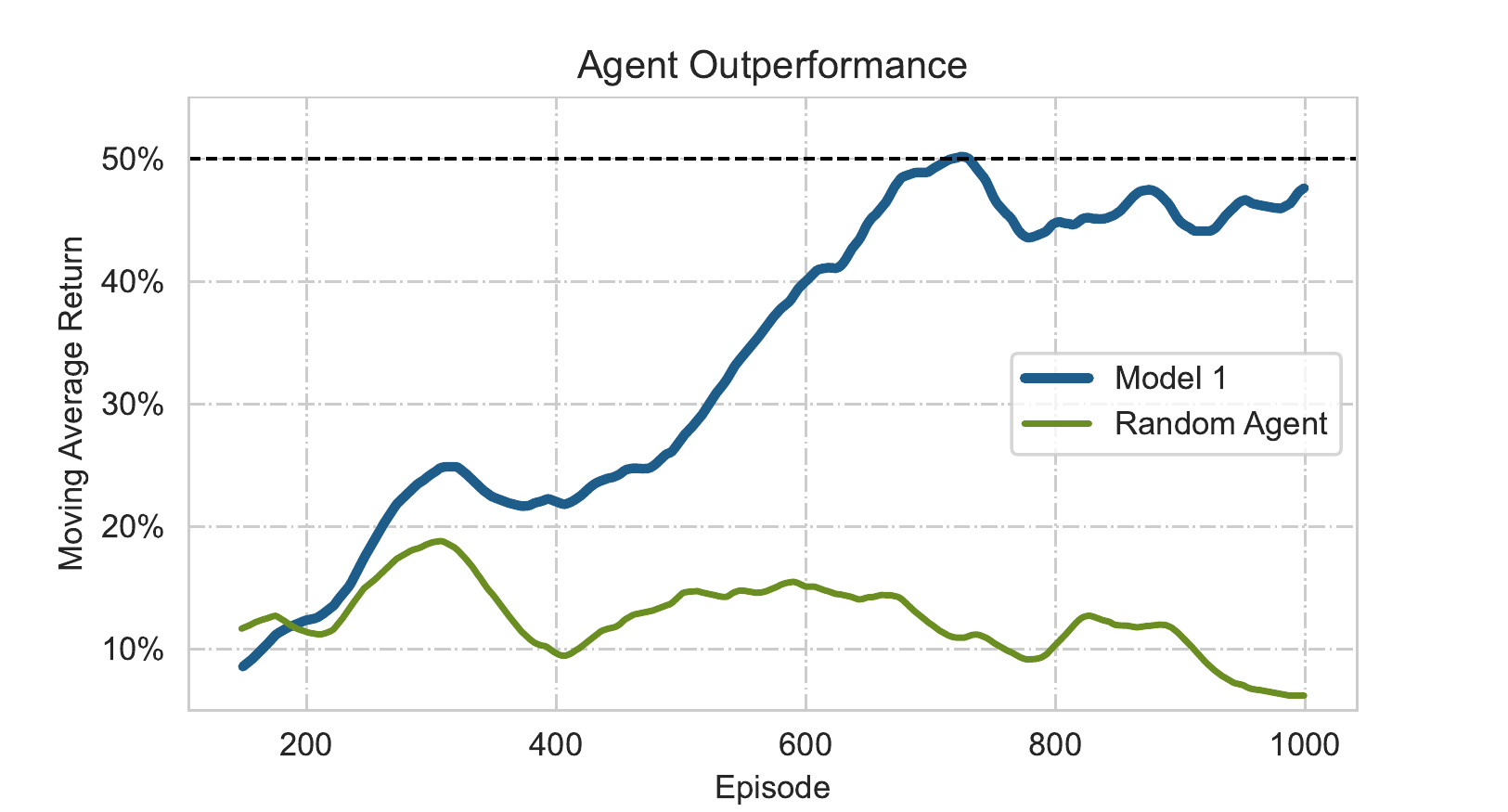}
     \end{subfigure}
    \caption{In-sample Performance of the trading agent in Model 1.}
\label{fig:1}
\end{figure}

In Figure \ref{fig:1} (left part), we display the training progress of our trading agent in Model 1 over 1000 episodes. It is apparent that the random actions through a high epsilon in the first 200 episodes ($\varepsilon$-greedy policy) allows the agent to explore but on the other hand, is coupled with a negative cumulative return. The return reverses toward the positive at an epsilon of less than 0.5. At the end of the training (1000 years), Model 1 has still not managed to exceed the market's return. The comparison between Model 1 and a model where actions are random (random agent) clearly reveals the progress of our trading agents through the episodes (Figure \ref{fig:1}). Outperformance can be achieved by the agent if it manages to achieve a higher net asset value than the market in 50\% of the cases over the last 50 episodes (moving average).

As for daily returns, the left part of Figure \ref{fig:2} clearly shows a shift triggered by transaction costs. The shift seems minor, but when accumulated over a year, it can have a significant impact on the NAV of the trading agent. Furthermore, daily returns excluding fees indicate a more right-skewed distribution.

\begin{figure}[ht]
     \centering
     \begin{subfigure}[b]{0.49\textwidth}
         \centering
         \includegraphics[width=\textwidth]{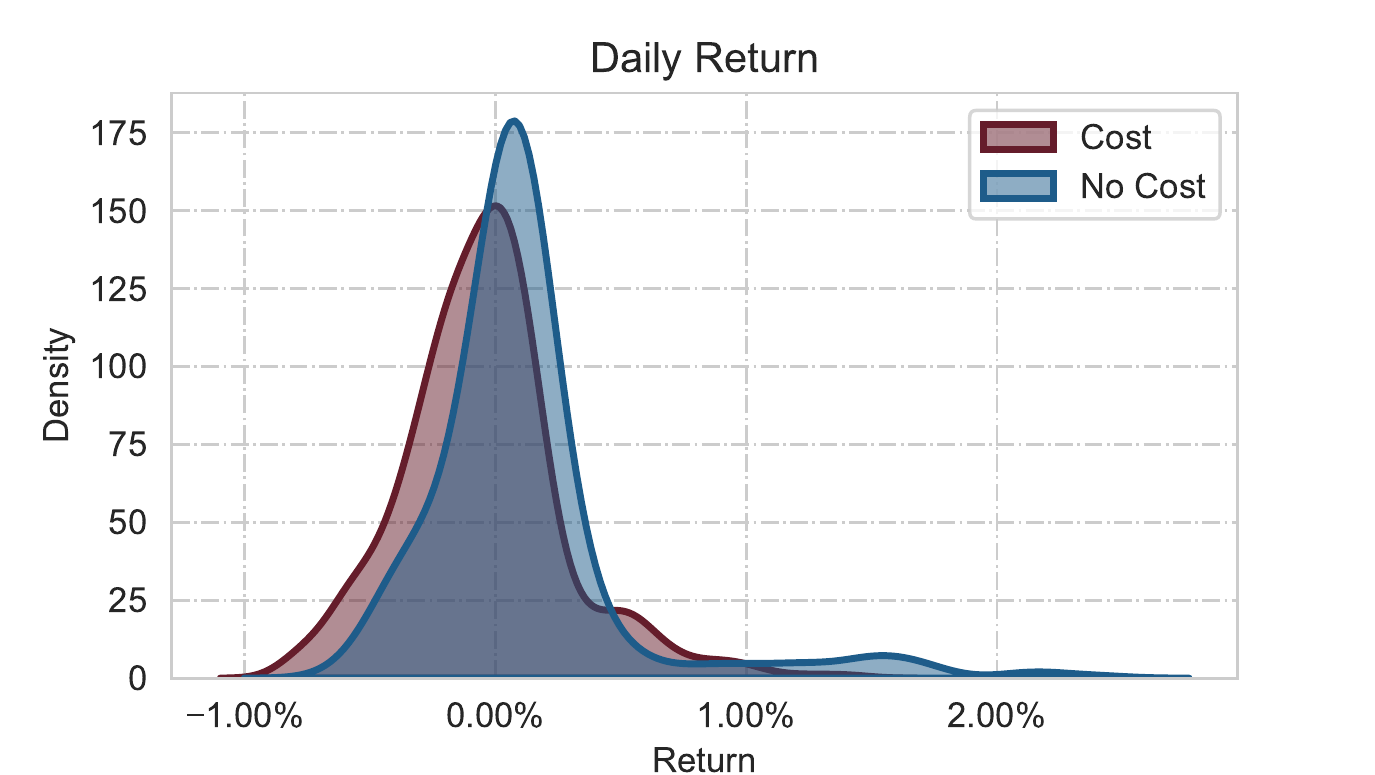}
     \end{subfigure}
     \begin{subfigure}[b]{0.49\textwidth}
         \centering
         \includegraphics[width=\textwidth]{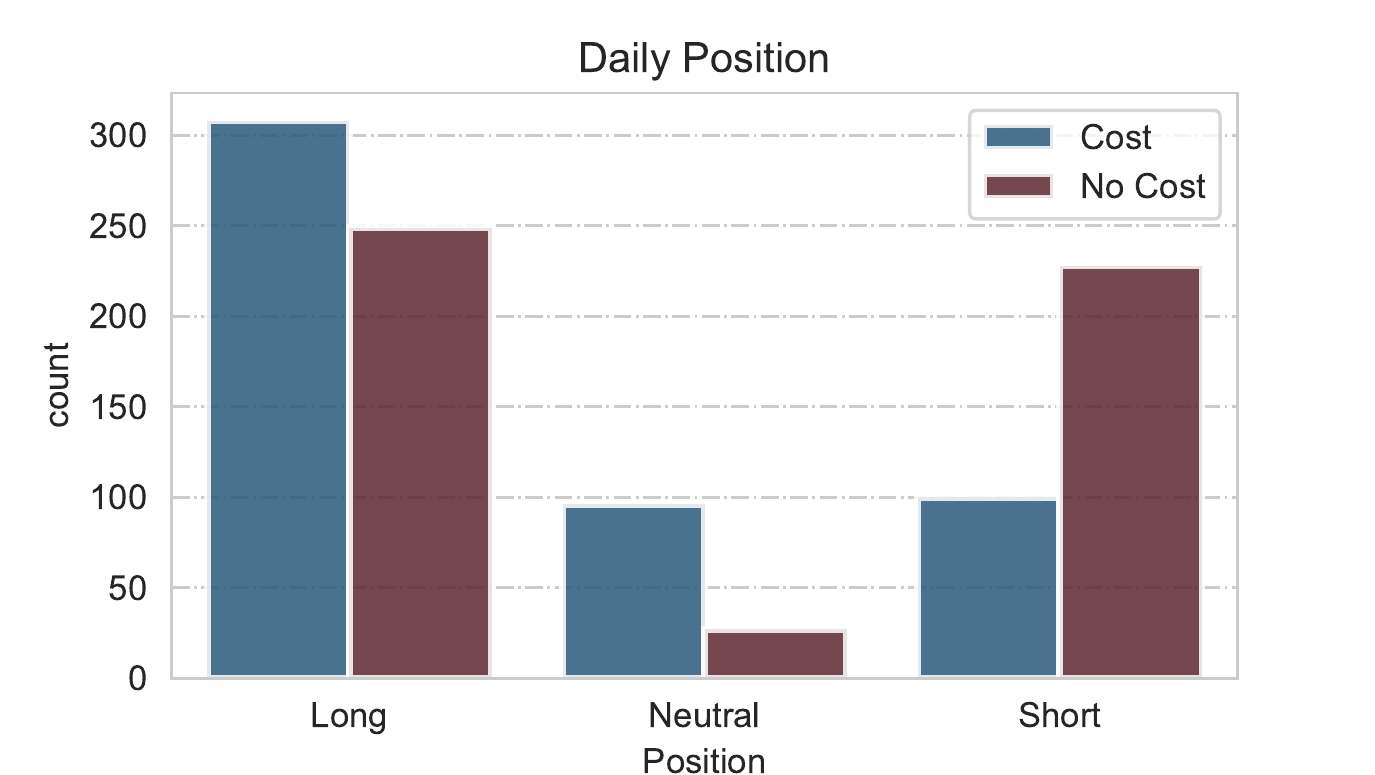}
     \end{subfigure}
    \caption{Out-of-sample Return and Position of the trading agent in Model 1.}
\label{fig:2}
\end{figure}

The right part of Figure \ref{fig:2} appears to be interesting. In a no-cost environment, the trading agent usually avoids being neutral, whereas when costs are included, the agent more often executes a neutral position. This makes sense, because to be profitable, the cost of changing a position must be less than the return on the new position. We must keep in mind that the change from long to short (and vice versa) involves double transaction costs.

We now move on to the comparison of all models, which can be considered the centerpiece of the results. The following figure \ref{fig:3} gives an overview of all models in a training and testing setup where regular trading costs are incurred.

\begin{figure}[ht]
     \centering
     \begin{subfigure}[b]{0.49\textwidth}
         \centering
         \includegraphics[width=\textwidth]{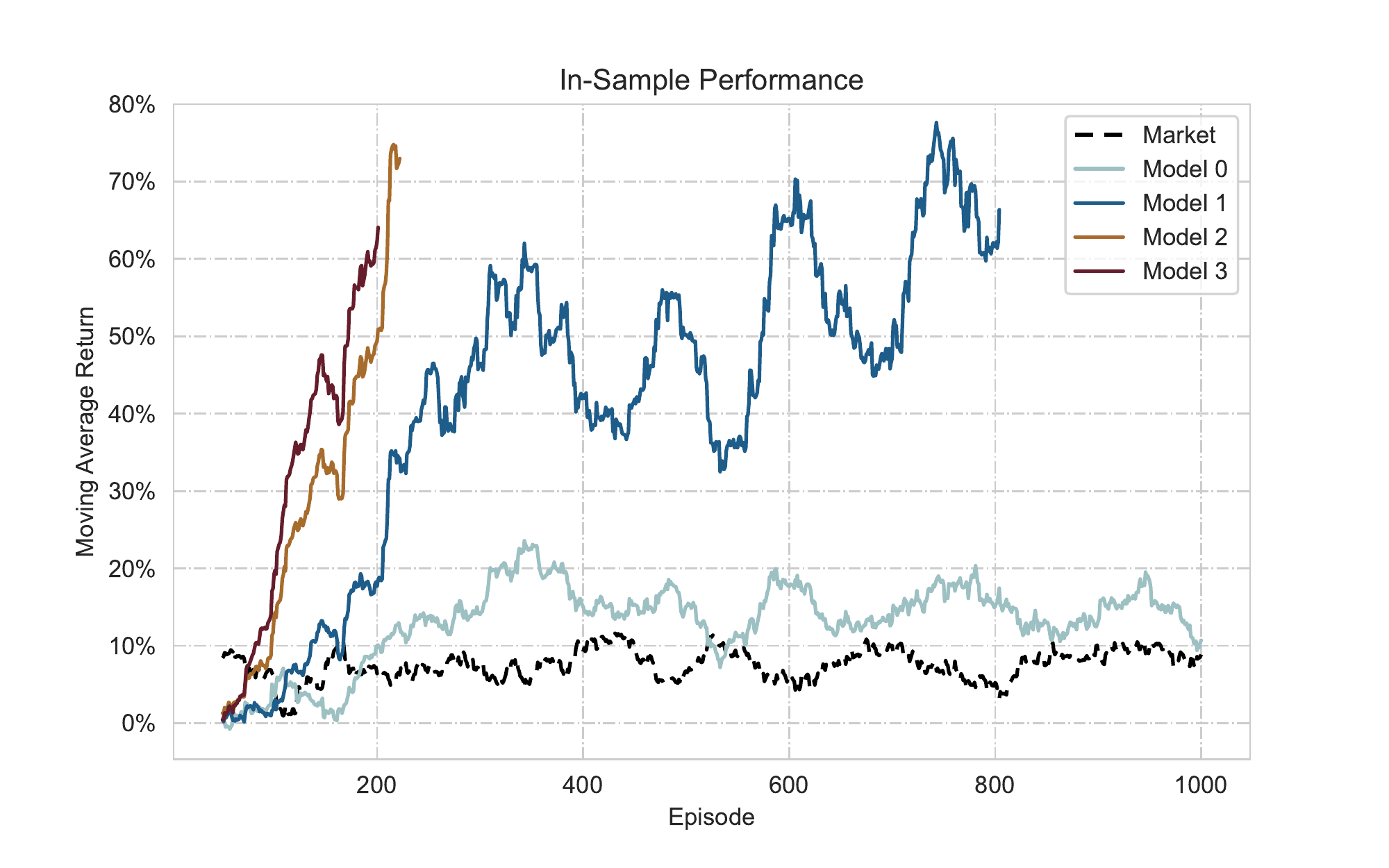}
     \end{subfigure}
     \begin{subfigure}[b]{0.49\textwidth}
         \centering
         \includegraphics[width=\textwidth]{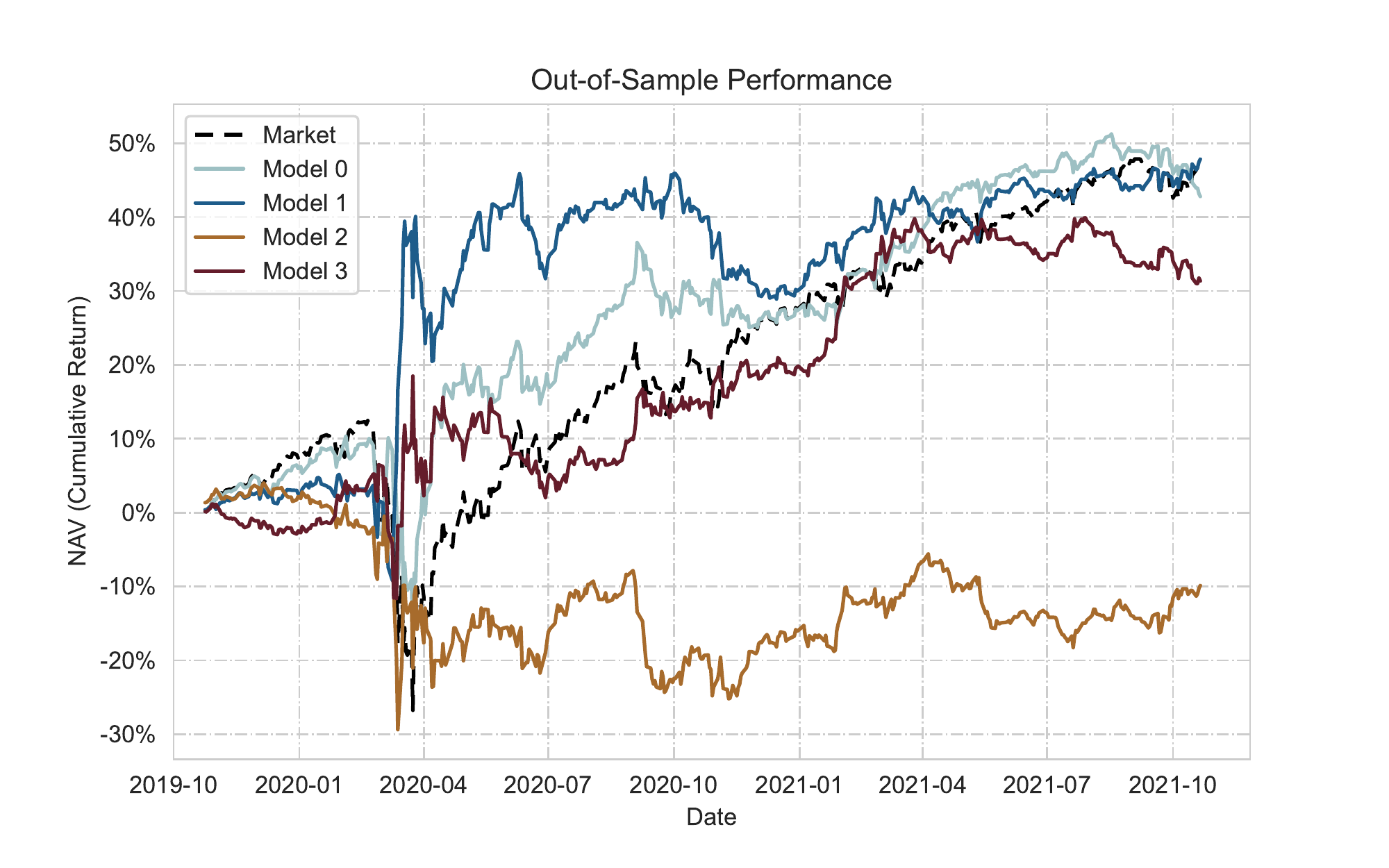}
     \end{subfigure}
    \caption{Performance of all models in-sample and out-of-sample in a no-cost environment.}
\label{fig:3}
\end{figure}

It is immediately noticeable that Model 2 and Model 3 perform significantly better in training (in-sample) than the other models and that their training duration is shorter. Model 3 finishes training after 200 episodes with a peak moving average return of 64\%. The training duration of Model 2 and 3 is shortened by the threshold of outperforming the market for 25 consecutive years, which counteracts overfitting. In contrast, if only the lagged returns of the S\&P 500 are used as the state space (Model 0), the trading agent behaves similarly to the market within 1000 episodes.

The out-of-sample performance shows a different result. The test period can be split into three major phases: before, during, and after the crisis, starting in March 2020. All models appear to perform identical pre-crisis, while especially Model 0 and Model 1 short the market during the March drop, resulting in a high net asset value (NAV). Given that Model 0 and Model 1 emerged as a winner from the crisis, they also were able to maintain the high net asset value over the next year. In addition, Model 3 recovered from a slight decline after the breakout and also kept its NAV on par with the other models. The trading agent in Model 2, on the other hand, failed to generate positive returns. It can be observed that Model 2 did not recover from the crisis, but kept its NAV constant afterwards. Surprisingly, only Model 0 managed to achieve a slightly higher NAV than the long-only benchmark (market).

The in-sample performance suggests that the models with higher features perform better, while the out-of-sample performance shows a contrasting behavior. In addition to the figures, we evaluate the results with performance keys in Table \ref{tab:perf}. The performance keys from the in-sample outcome can be found in the appendix section \ref{app:7.3} and \ref{app:7.4}. The following metrics are used to evaluate the out-of-sample results:

\begin{itemize}
    \itemsep0em
    \item $\text{E(R)}$: annualized expectation of daily return
    \item $\text{std(R)}$: annualized standard deviation of daily return
    \item Sharpe: annualized Sharpe Ratio 
    $E(R)/(std(R))$
    \item MSE: Mean-squared-error of the actions taken by the agent $1/n \sum_{i=1}^{n}(A_i - \hat{A_i})$
    \item Accuracy Score: Accuracy score of the actions taken by the agent $
    correct\ predictions/ all\ predictions$
\end{itemize}

To estimate the MSE and Accuracy, the optimal actions were determined by the out-of-sample time series. This estimation was necessary because our models include costs and thus allow for profitable neutral positions, rather than just assigning positive and negative returns to long and short positions.

\begin{table}
\renewcommand{\arraystretch}{1.5}
\centering
\caption{Out-of-Sample Performance Keys}
\begin{tabular}{| l | r | r | r | r | r |}
    \hline
                & E(R)      & std(R)    & Sharpe    & MSE       & Accuracy  \\\hline
    Market      & 22.32\%   & 25.26\%   & 0.883     & 1.680     & 0.568     \\
    Model 0     & 19.77\%   & 25.18\%   & 0.785     & 1.654     & 0.540     \\
    Model 1     & 22.87\%   & 25.0\%    & 0.915     & 1.758     & 0.490     \\
    Model 2     & -5.47\%   & 24.97\%   & -0.219    & 1.848     & 0.478     \\
    Model 3     & 15.91\%   & 23.53\%   & 0.676     & 1.784     & 0.458     \\\hline
\end{tabular}
\label{tab:perf}
\end{table}

From the Table \ref{tab:perf} we can deduce that all models expect Model 2 and the market have generated a positive return. It is also evident that all models have similar volatility. In terms of expected return or Sharpe ratio, Model 1 performs strongest and outperforms the benchmark. Another remarkable fact emerges from the accuracy scores. Although Model 1 performed better, the market achieved higher accuracy. In addition, Models 1-3 did not pass the 50\% threshold, but can still be considered adequate. For comparison, a trading agent executing random actions would achieve a success rate of about 33\%. 

Overall, it can be concluded that the simpler models emerged as the winner and the model with higher input did not show any significant performance improvement regarding the out-of-sample. However, we have included the results of all models with the other environment settings in the appendix. They demonstrate that models with higher features perform different in a crisis-free environment, or that trading costs can reduce the performance of the trading agent.

\section{Conclusion and Outlook}
This paper presents key applications for Double Deep Q-learning with finance. We adopt the DDQN framework with its environment \cite{jansen2020machinelearning} to build four models for trading the S\&P 500. We start with the simplest model with the lagged S\&P 500 itself as a feature and gradually add additional features from other assets to the models. Furthermore, the models are trained and tested in a cost/no-cost environment and pre- and post-crisis. So, in total, we are referring to 16 models. We compare all four models with their specific settings (cost/no-cost, pre/post crisis) and furthermore take the market with its simple long-and-hold strategy as a benchmark. We evaluate in-sample and, more importantly, out-of-sample results using performance indicators such as the NAV or Sharpe ratio. 

The result of this study shows that the agent can achieve strong in-sample performance and, under certain circumstances, is able to transfer these results when being out-of-sample. Each model is able to outperform the market in-sample with a given episode length. It is also characteristic that this outperformance occurs more quickly when additional features are added to the model. As we move out of the sample, we find that almost all models are capable of producing reasonable results depending on the environment. Moreover, the models accounted for the various environmental factors, such as trade costs. The integration of trade costs made the neutral position more favorable in certain cases, which the trade agent took advantage of. However, we can also conclude that the performance of all models drops when costs are involved, especially when they are greatly increased.

There are some general remarks about the design of this study that we would like to mention. Due to the training length of 1000 episodes per model, it was quite difficult to test each model multiple times, so uncertainties could not be avoided and the results may have been less robust. In addition, the models, especially those with higher features, tend to overfit. We must keep in mind that trading a single asset strongly increases the risk of overfitting. Another crucial part is the setting of the hyperparameters. This is especially important in an environment where we are exposed to a lot of noise. Unfortunately, there is no facutal basis for applications of RL in finance. One can set the parameters based on other work or experiments, but it has proven difficult to find the optimal set.
 
Nonetheless, we were able to demonstrate that the trading agent is certainly capable of performing proficiently. For further research, we suggest the following improvements:

\begin{enumerate}[nosep]
    \item{Extending the target from trading a single asset to selecting from multiple assets. Appropriate asset allocation \cite{2102.06233} reduces volatility while maintaining returns. This also helps to counteract overfitting when training the model.} 
    \item{Evaluating and extending the feature set. Choosing lags based on autocorrelation, adding new features such as momentum indicators, and more will help the model optimize more accurately \cite{1502.01073}.}
    \item{Manage risks and seize the positions. In our experiment, all positions had the same weighting. Adjusting this weighting to occurrences like bear or bull markets increases the NAV. In addition, the introduction of certain thresholds as an emergency exit option can help avoid drawdowns \cite{2201.09058}.}
\end{enumerate}

In conclusion, we have illustrated the application of a simple trading agent that uses a small number of features. Reinforcement Learning, especially in combination with neural networks, is often considered as the most promising approach for algorithmic trading. Our study has shown that the application of a DDQN algorithm can be an appropriate choice within a financial framework. Nevertheless, creating a more realistic, complicated environment that replicates the tasks of a real investor can be challenging. The current growing interest in machine learning and algorithmic trading may lead to large-scale experiments that can produce new findings. An exciting complementary approach is Inverse Reinforcement Learning \cite{1806.06877}, which attempts to identify the agent's reward function given  its behavior.

\section{Bibliography}

\bibliographystyle{unsrt}  
\bibliography{main}

\clearpage

\section{Appendix}
\subsection{Data}

\begin{figure}[ht]
    \centering
    \includegraphics[width=\textwidth]{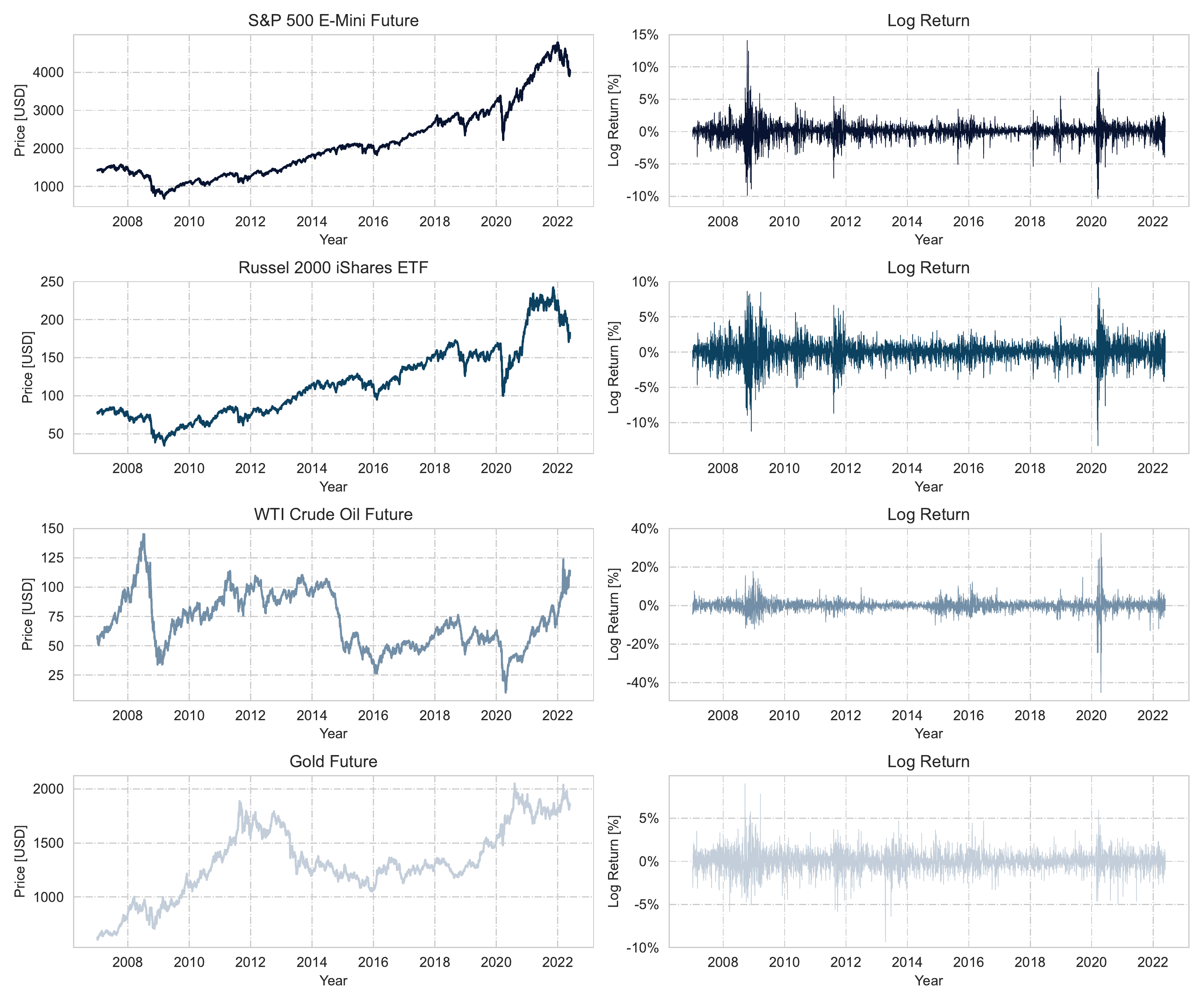}
    \caption{Asset Price Movement and Log Return.}
\end{figure}

\begin{table}
\renewcommand{\arraystretch}{1.5}
\centering
\caption{Asset Performance Keys of Log Returns (Annual)}
\begin{tabular}{| l | r | r | r | r | r |}
    \hline
                & E(R)      & std(R)    & Sharpe    & Skewness  & Kurtosis  \\\hline
    S\&P 500     & 8.94\%    & 20.65\%   & 0.433     & -0.135    & 14.602    \\
    Russel 2000 & 8.75\%    & 25.35\%   & 0.345     & -0.410    & 6.555     \\
    WTI         & 15.02\%   & 45.73\%   & 0.328     & -0.055    & 34.762    \\
    Gold        & 8.64\%    & 17.98\%   & 0.481     & -0.135    & 5.696     \\\hline
\end{tabular}
\end{table}

\clearpage

\begin{figure}[ht]
    \centering
    \includegraphics[width=0.50\textwidth]{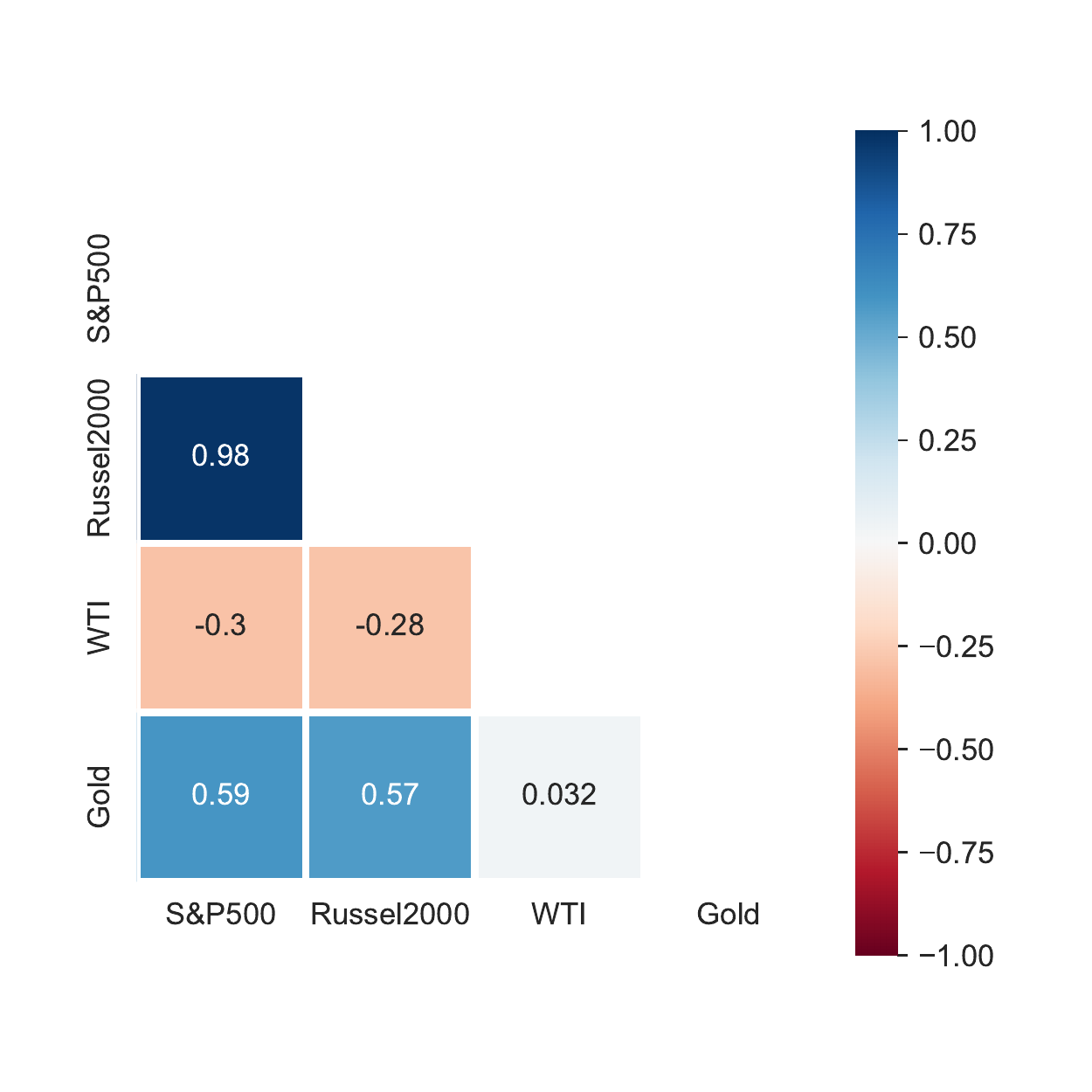}
    \caption{Correlation between assets.}
\end{figure}

\subsection{Hyperparameters}\label{hyper}

This section describes all hyperparameters used in the experiment. The hyperparameters can be divided into general and DDQN-specific parameters:

\begin{itemize}
    \item {\bf Epsiodes:} This refers to the fact that the agent goes through one cycle, i.e. one financial year. It is referred as $M$ in the DDQN algorithm \ref{alg:cap}.
    \item {\bf Episode length:} The episode length can be considered as the trading days. In our case, there are 252 days of trading. This is denoted as $T$ in the DDQN algorithm \ref{alg:cap}.
    \item {\bf Trading cost:} The costs incurred each time the position is changed. In our experiment, we used one basis point per trade, or 1 bps.
    \item {\bf Time cost:} The costs incurred if no change in position is made. This value is set to 0.1 bps.
\end{itemize}

\begin{itemize}

    \item {\bf Discount factor:} The factor for discounting future earnings, which can also be found in Section \ref{agent_policy}.
    \item {\bf Learning rate:} The step size for each iteration in optimizing the Q-learning function, or the weighting of adding the new estimate to the current one \ref{eq:3.7}. 
    \item {\bf Exploration probability:} This is described as the probability $\varepsilon$ in a $\varepsilon$-greedy policy of choosing a random action over the optimal action.
    \item {\bf Update target network:} The number of episodes that elapse before the target network is updated with the online network weights. It is denoted as $\tau$ in the DDQN algorithm \ref{alg:cap}.
    \item {\bf Replay capacity:} The capacity with which the replay memory can store tuples $(s_t, a_t, r_t, s_{t+1})$.
    \item {\bf Batch size:} The number of tuples $e_t = (s_t, a_t, r_t, s_{t+1})$ passed to the DDQN as experience replay. Also called mini-batch.
    \item {\bf Optimizer:} This is the optimization algorithm that is passed to the DDQN online network. We use the adam optimizer \cite{2206.02034}.
\end{itemize}

\clearpage

\subsection{Results Pre-Crisis} \label{app:7.3}

\begin{figure}[ht]
     \centering
     \begin{subfigure}[b]{0.49\textwidth}
         \centering
         \includegraphics[width=\textwidth]{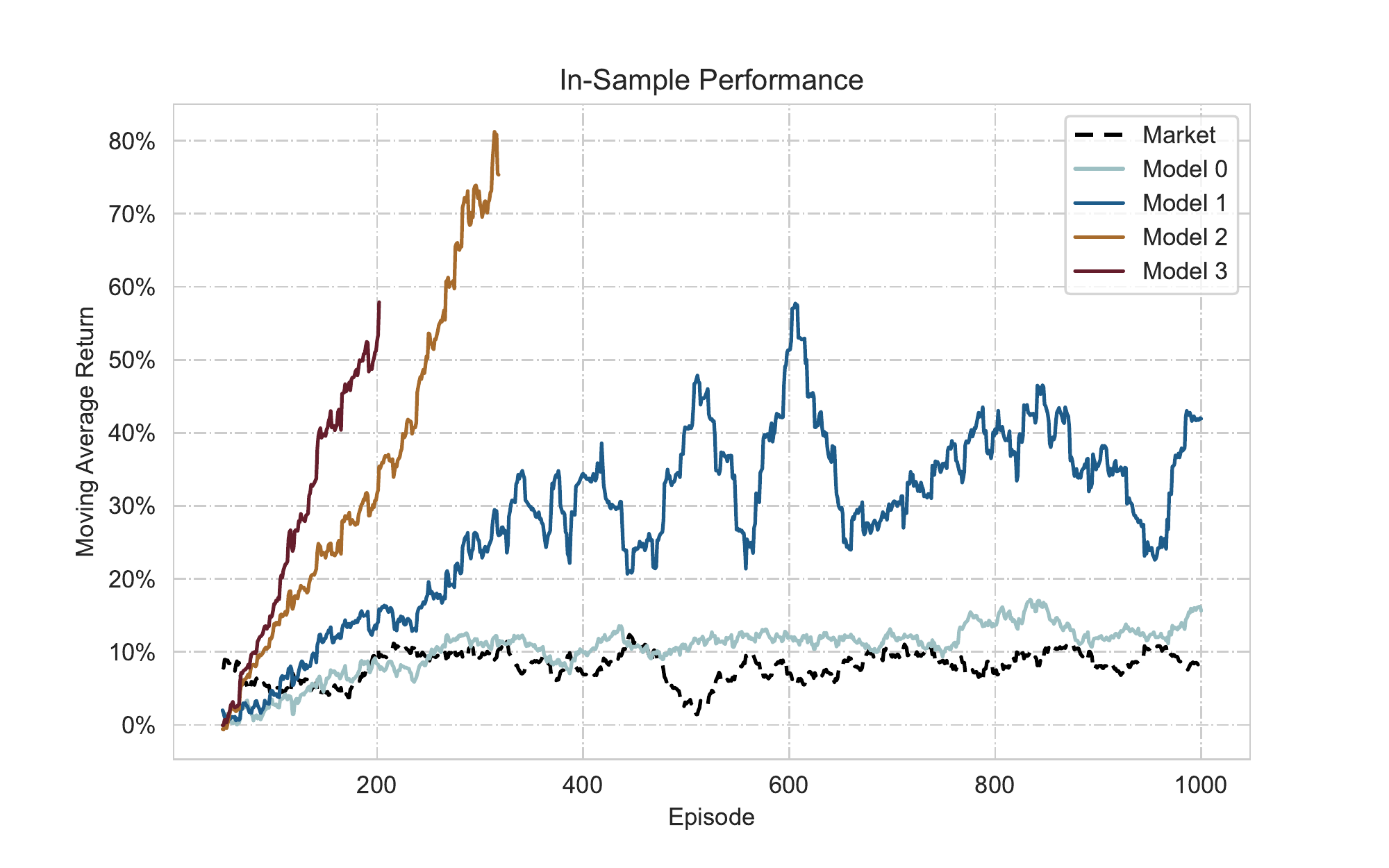}
     \end{subfigure}
     \begin{subfigure}[b]{0.49\textwidth}
         \centering
         \includegraphics[width=\textwidth]{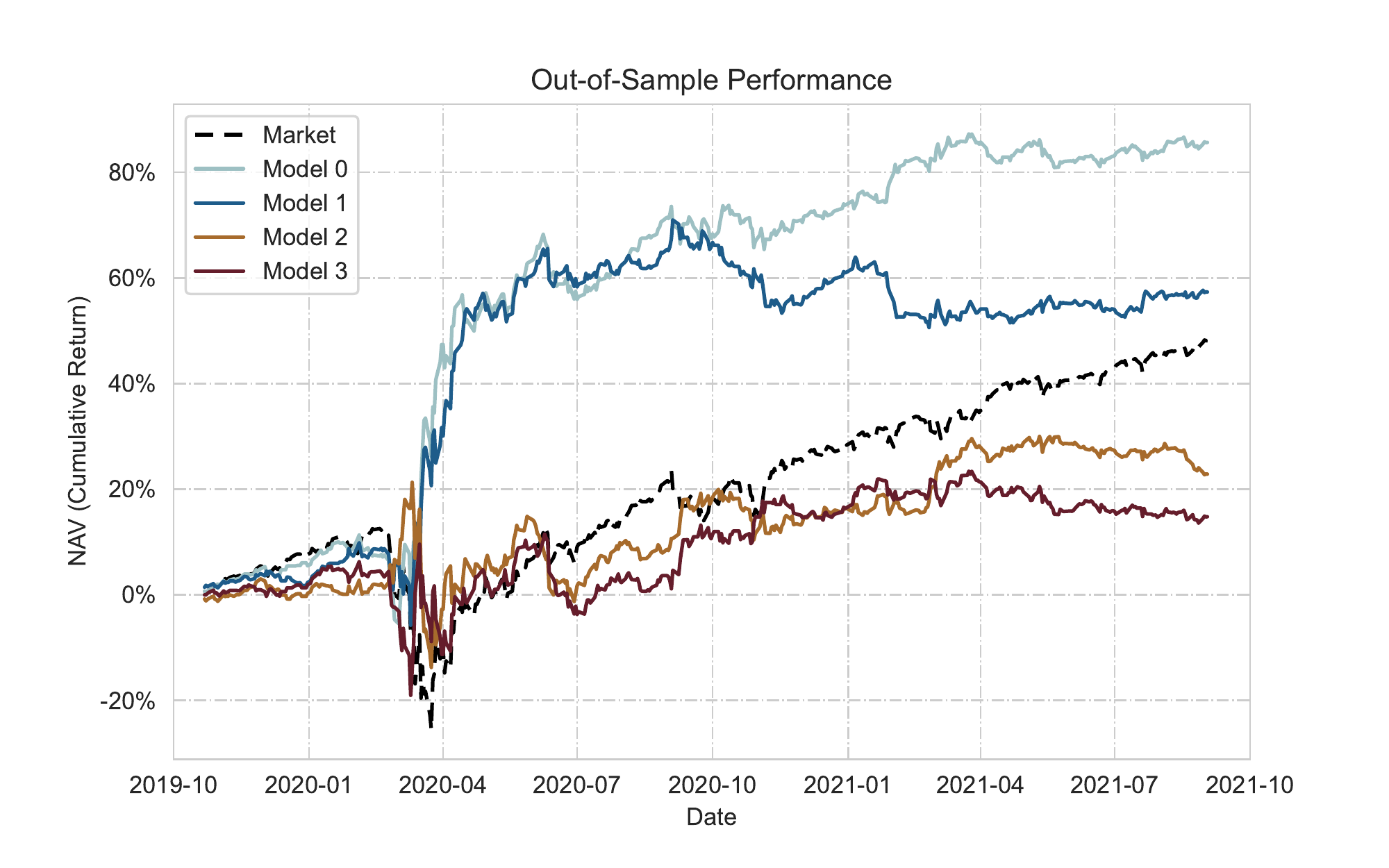}
     \end{subfigure}
    \caption{Performance of all models in-sample and out-of-sample {\bf without costs}.}
\end{figure}

\begin{table}
\scalebox{0.87}{
    \begin{minipage}{.49\linewidth}
        \renewcommand{\arraystretch}{1.5}
        \centering
        \caption{In-Sample Performance Keys}
        \begin{tabular}{| l | r | r | r |}
        \hline
                    & E(R)      & std(R)    & Sharpe    \\\hline
        Market      & 8.04\%    & 14.69\%   & 0.547     \\
        Model 0     & 10.46\%   & 15.07\%   & 0.694     \\
        Model 1     & 28.07\%   & 45.09\%   & 0.622     \\
        Model 2     & 34.59\%   & 50.71\%   & 0.682     \\
        Model 3     & 28.62\%   & 42.71\%   & 0.670     \\\hline
    \end{tabular}
    \end{minipage}
    }
\scalebox{0.87}{
    \begin{minipage}{.49\linewidth}
        \renewcommand{\arraystretch}{1.5}
        \centering
        \caption{Out-of-Sample Performance Keys}
        \begin{tabular}{| l | r | r | r | r | r |}
        \hline
                    & E(R)      & std(R)    & Sharpe    & MSE       & Accuracy  \\\hline
        Market      & 24.68\%   & 26.35\%   & 0.937     & 1.650     & 0.581     \\
        Model 0     & 44.88\%   & 26.17\%   & 1.715     & 1.652     & 0.560     \\
        Model 1     & 28.49\%   & 26.23\%   & 1.086     & 1.795     & 0.513     \\
        Model 2     & 12.36\%   & 25.47\%   & 0.485     & 1.829     & 0.498     \\
        Model 3     & 7.6\%     & 25.92\%   & 0.293     & 1.842     & 0.459     \\\hline
    \end{tabular}
    \end{minipage}
    }
\end{table}

\begin{figure}[ht]
     \centering
     \begin{subfigure}[b]{0.49\textwidth}
         \centering
         \includegraphics[width=\textwidth]{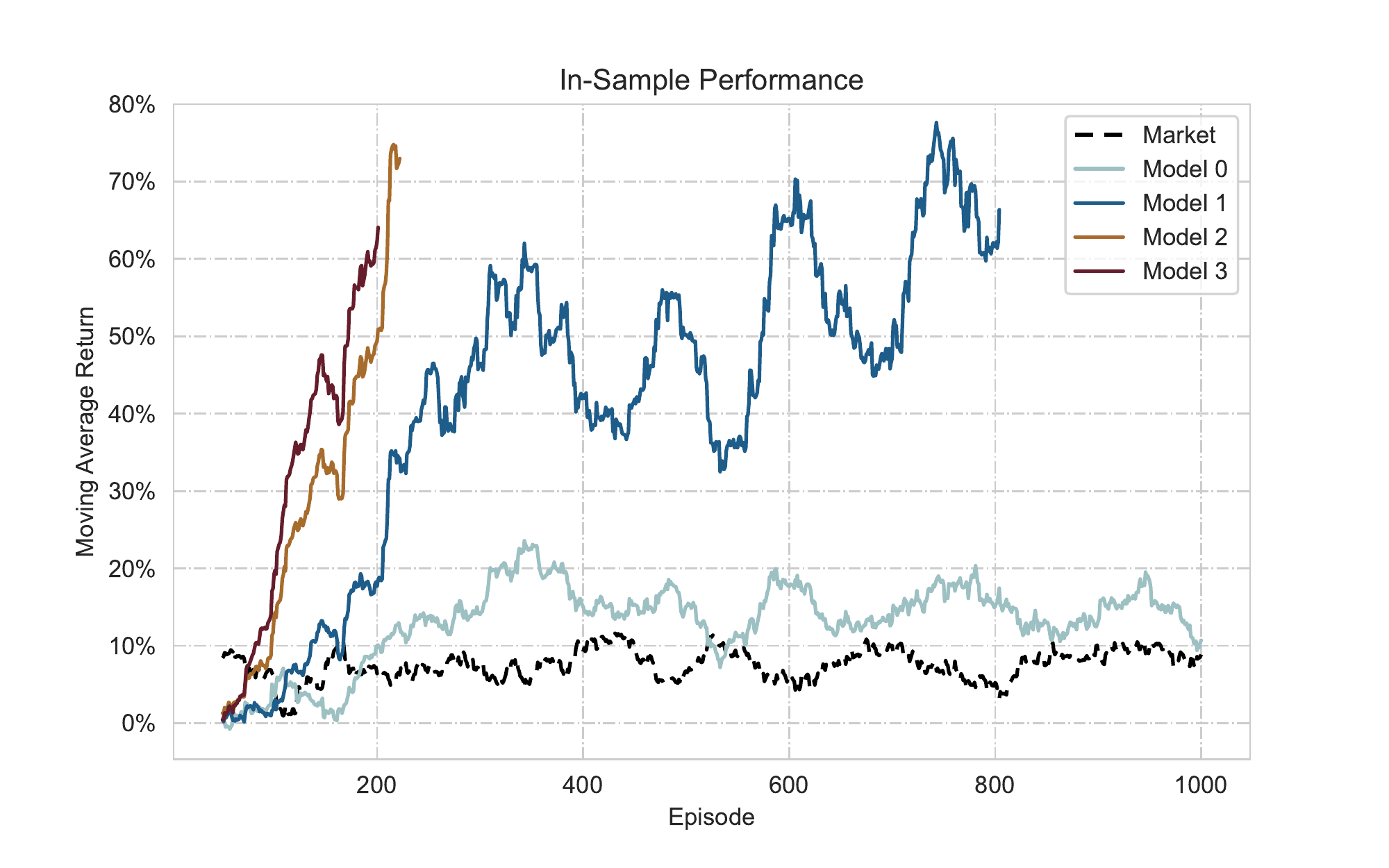}
     \end{subfigure}
     \begin{subfigure}[b]{0.49\textwidth}
         \centering
         \includegraphics[width=\textwidth]{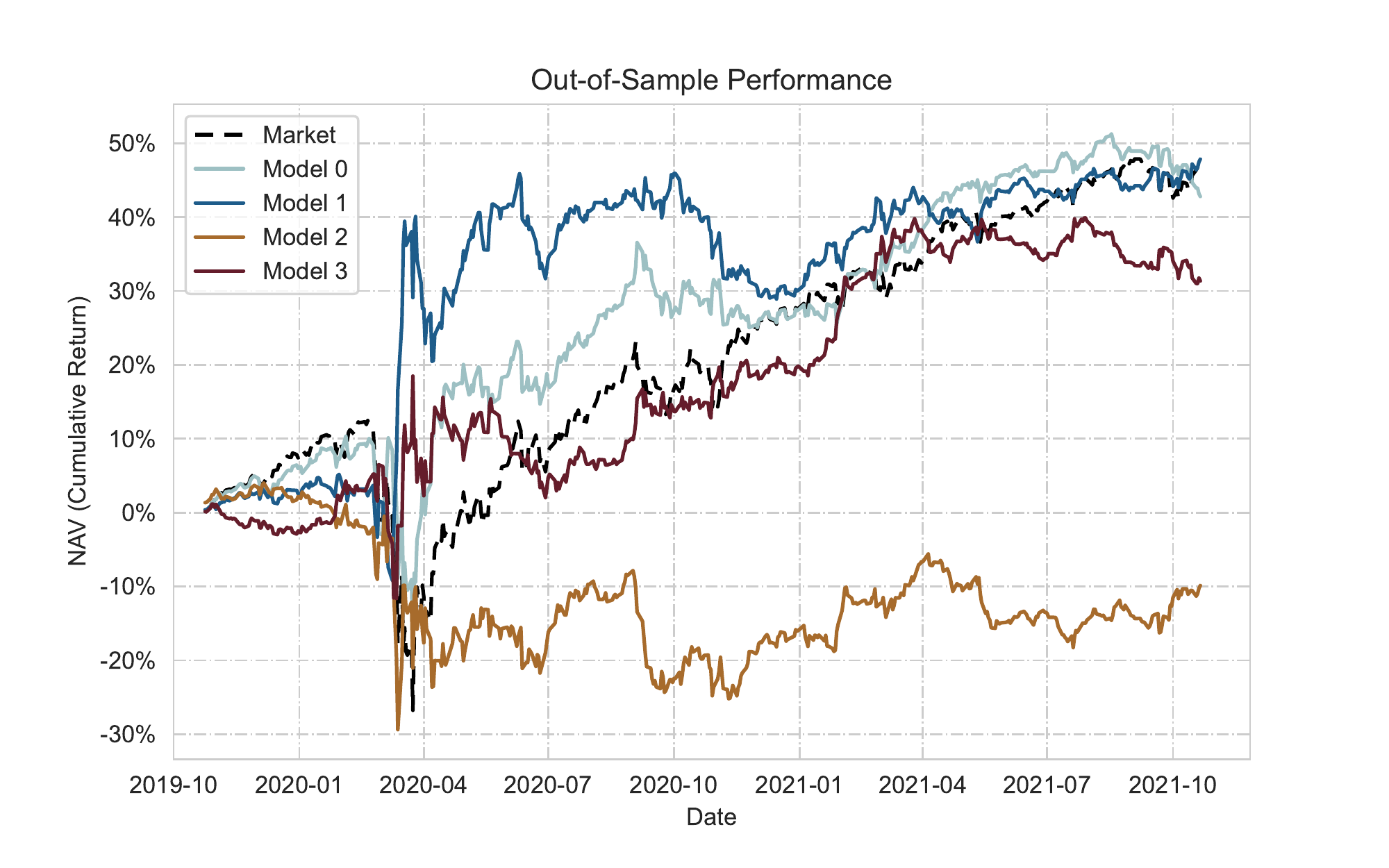}
     \end{subfigure}
    \caption{Performance of all models in-sample and out-of-sample {\bf with costs}.}
\end{figure}

\begin{table}
\scalebox{0.87}{
    \begin{minipage}{.49\linewidth}
        \renewcommand{\arraystretch}{1.5}
        \centering
        \caption{In-Sample Performance Keys}
        \begin{tabular}{| l | r | r | r |}
        \hline
                    & E(R)      & std(R)    & Sharpe    \\\hline
        Market      & 7.66\%    & 16.12\%   & 0.475     \\
        Model 0     & 12.78\%   & 21.0\%    & 0.609     \\
        Model 1     & 41.27\%   & 54.63\%   & 0.755     \\
        Model 2     & 31.8\%    & 51.92\%   & 0.612     \\
        Model 3     & 31.96\%   & 43.87\%   & 0.729     \\\hline
    \end{tabular}
    \end{minipage}
    }
\scalebox{0.87}{
    \begin{minipage}{.49\linewidth}
        \renewcommand{\arraystretch}{1.5}
        \centering
        \caption{Out-of-Sample Performance Keys}
        \begin{tabular}{| l | r | r | r | r | r |}
        \hline
                    & E(R)      & std(R)    & Sharpe    & MSE       & Accuracy  \\\hline
        Market      & 22.32\%   & 25.26\%   & 0.883     & 1.680     & 0.568     \\
        Model 0     & 19.77\%   & 25.18\%   & 0.785     & 1.654     & 0.540     \\
        Model 1     & 22.87\%   & 25.0\%    & 0.915     & 1.758     & 0.490     \\
        Model 2     & -5.47\%   & 24.97\%   & -0.219    & 1.848     & 0.478     \\
        Model 3     & 15.91\%   & 23.53\%   & 0.676     & 1.784     & 0.458     \\\hline
    \end{tabular}
    \end{minipage}
    }
\end{table}

\clearpage

\subsection{Results Post-Crisis} \label{app:7.4}

\begin{figure}[ht]
     \centering
     \begin{subfigure}[b]{0.49\textwidth}
         \centering
         \includegraphics[width=\textwidth]{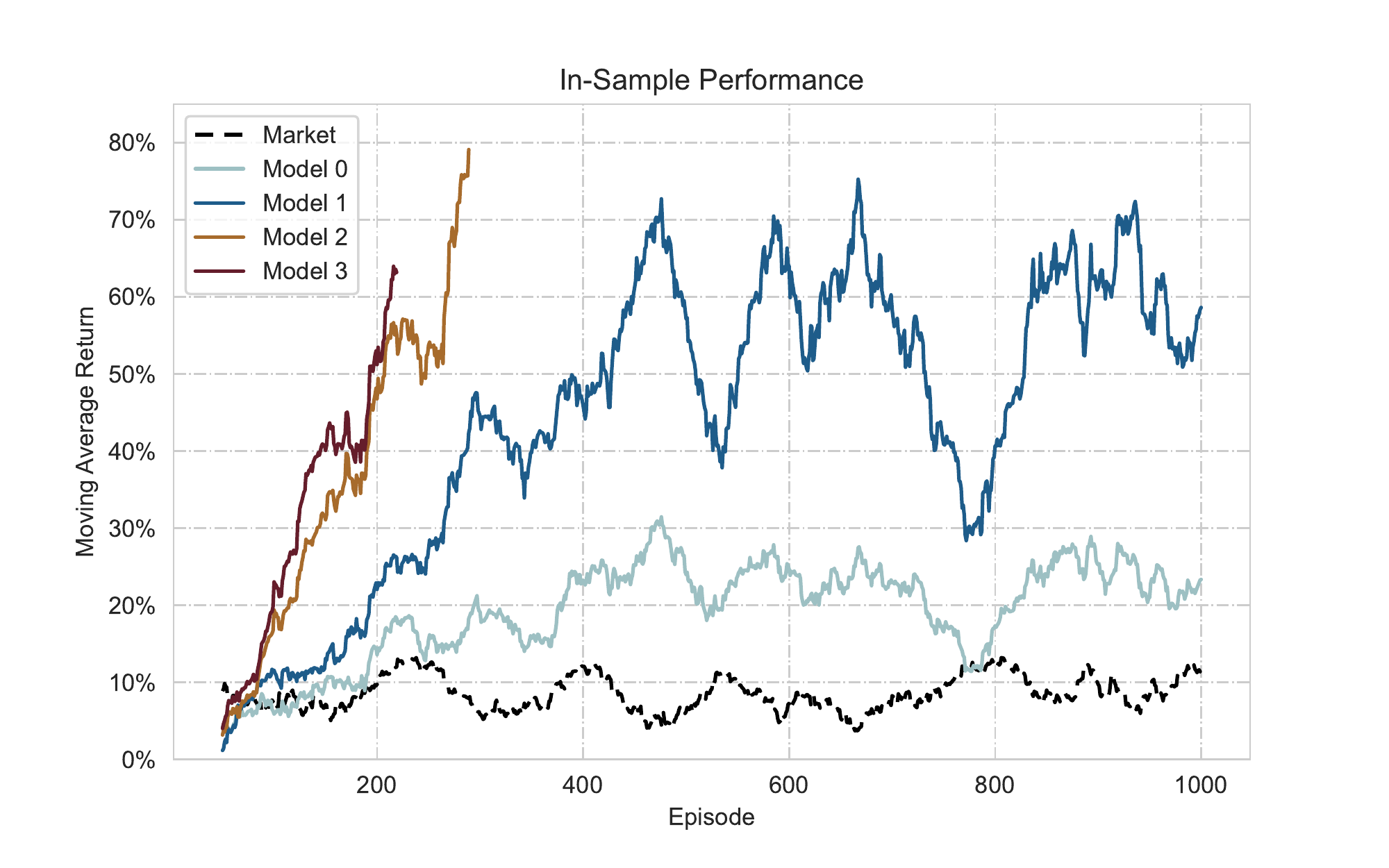}
     \end{subfigure}
     \begin{subfigure}[b]{0.49\textwidth}
         \centering
         \includegraphics[width=\textwidth]{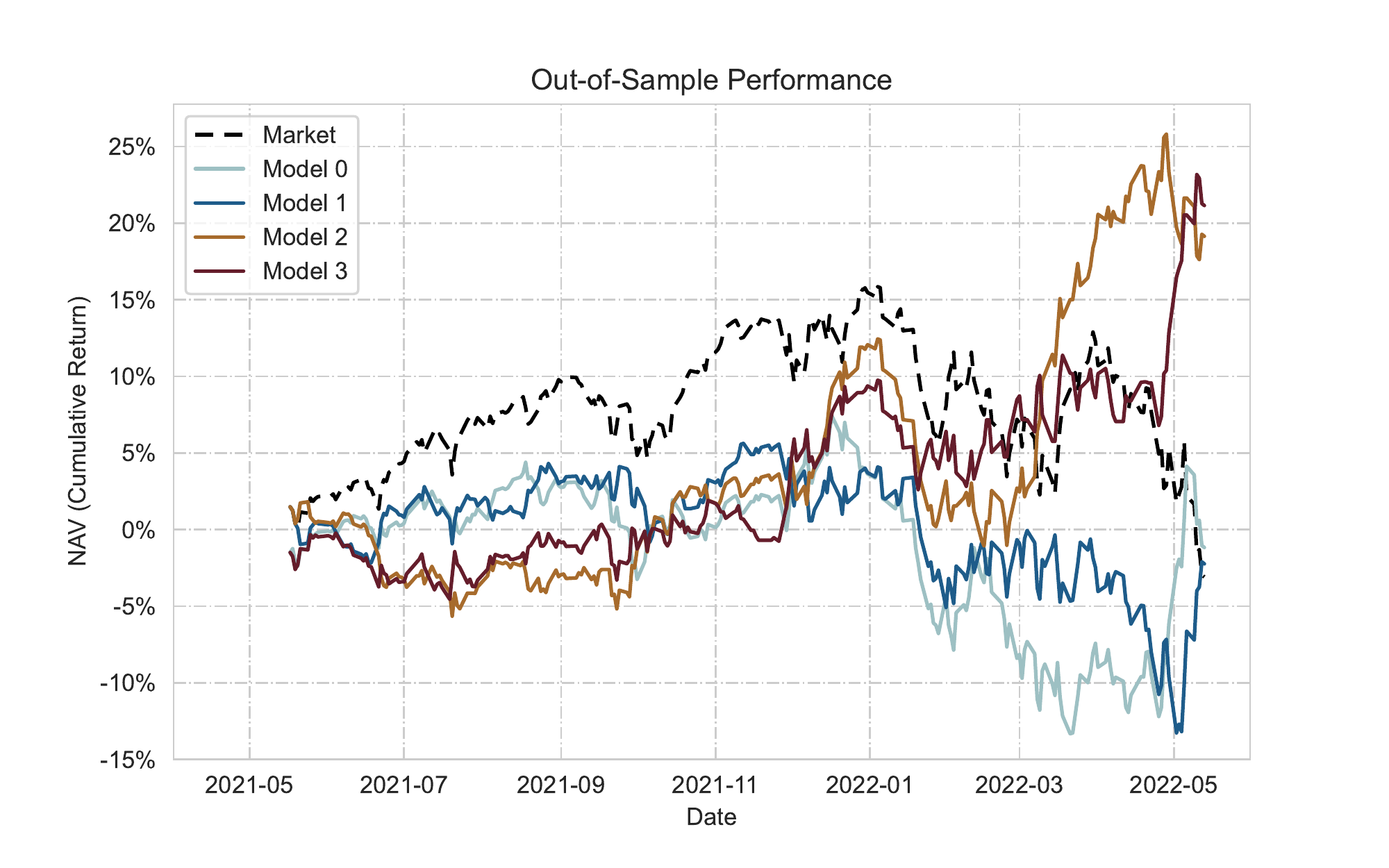}
     \end{subfigure}
    \caption{Performance of all models in-sample and out-of-sample {\bf without costs}.}
\end{figure}

\begin{table}
\scalebox{0.87}{
    \begin{minipage}{.49\linewidth}
        \renewcommand{\arraystretch}{1.5}
        \centering
        \caption{In-Sample Performance Keys}
        \begin{tabular}{| l | r | r | r |}
        \hline
                    & E(R)      & std(R)    & Sharpe    \\\hline
        Market      & 8.65\%    & 15.35\%   & 0.563     \\
        Model 0     & 18.93\%   & 24.09\%   & 0.786     \\
        Model 1     & 43.8\%    & 49.13\%   & 0.892     \\
        Model 2     & 37.55\%   & 49.5\%    & 0.759     \\
        Model 3     & 32.83\%   & 42.57\%   & 0.771     \\\hline
    \end{tabular}
    \end{minipage}
    }
\scalebox{0.87}{
    \begin{minipage}{.49\linewidth}
        \renewcommand{\arraystretch}{1.5}
        \centering
        \caption{Out-of-Sample Performance Keys}
        \begin{tabular}{| l | r | r | r | r | r |}
        \hline
                    & E(R)      & std(R)    & Sharpe    & MSE       & Accuracy  \\\hline
        Market      & -3.09\%   & 17.01\%   & -0.181    & 1.897     & 0.520     \\
        Model 0     & -1.17\%   & 17.01\%   & -0.069    & 1.992     & 0.496     \\
        Model 1     & -2.23\%   & 16.99\%   & -0.131    & 2.008     & 0.480     \\
        Model 2     & 19.14\%   & 16.36\%   & 1.170     & 2.000     & 0.464     \\
        Model 3     & 21.15\%   & 16.05\%   & 1.317     & 1.885     & 0.460     \\\hline
    \end{tabular}
    \end{minipage}
    }
\end{table}

\begin{figure}[ht]
     \centering
     \begin{subfigure}[b]{0.49\textwidth}
         \centering
         \includegraphics[width=\textwidth]{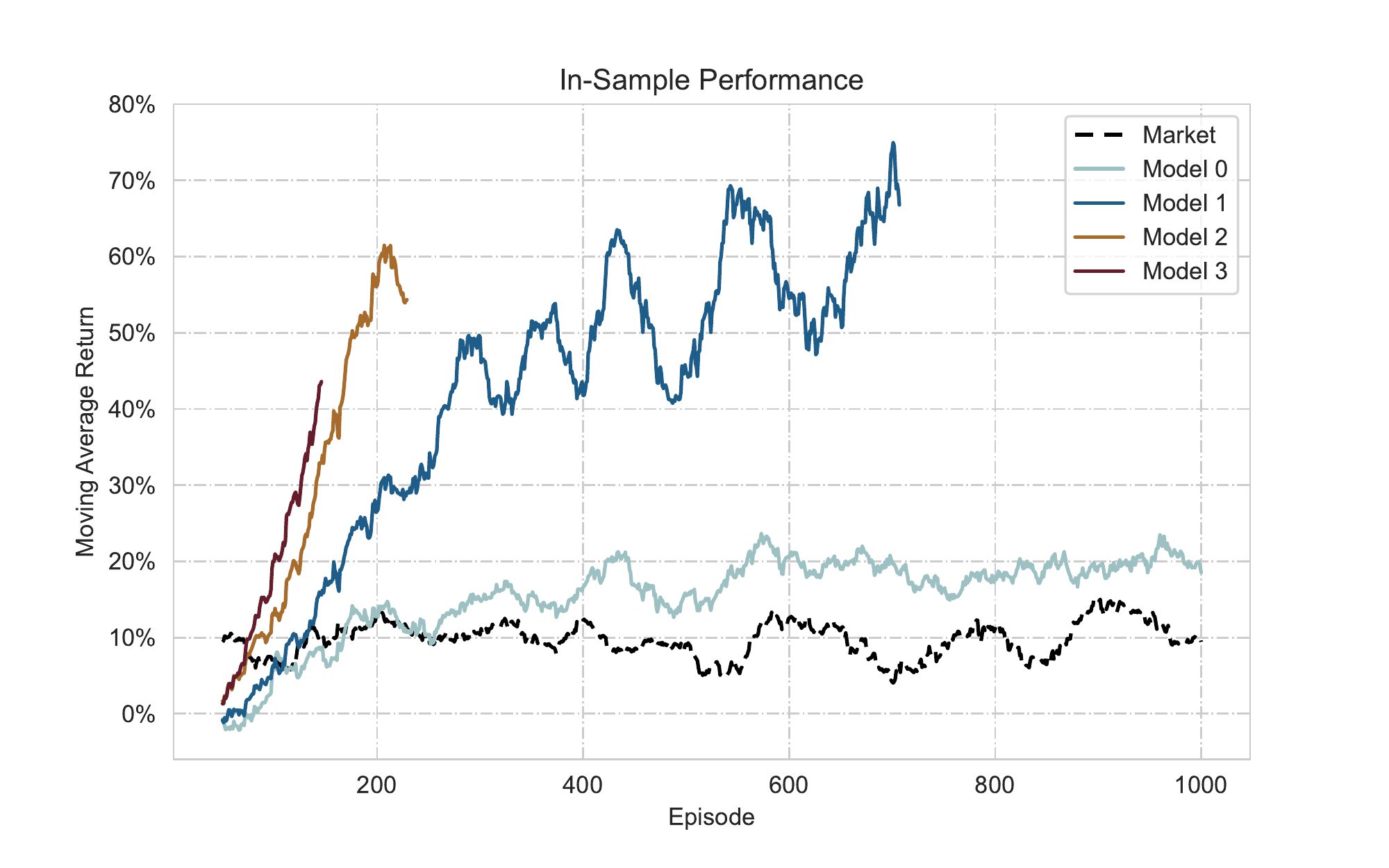}
     \end{subfigure}
     \begin{subfigure}[b]{0.49\textwidth}
         \centering
         \includegraphics[width=\textwidth]{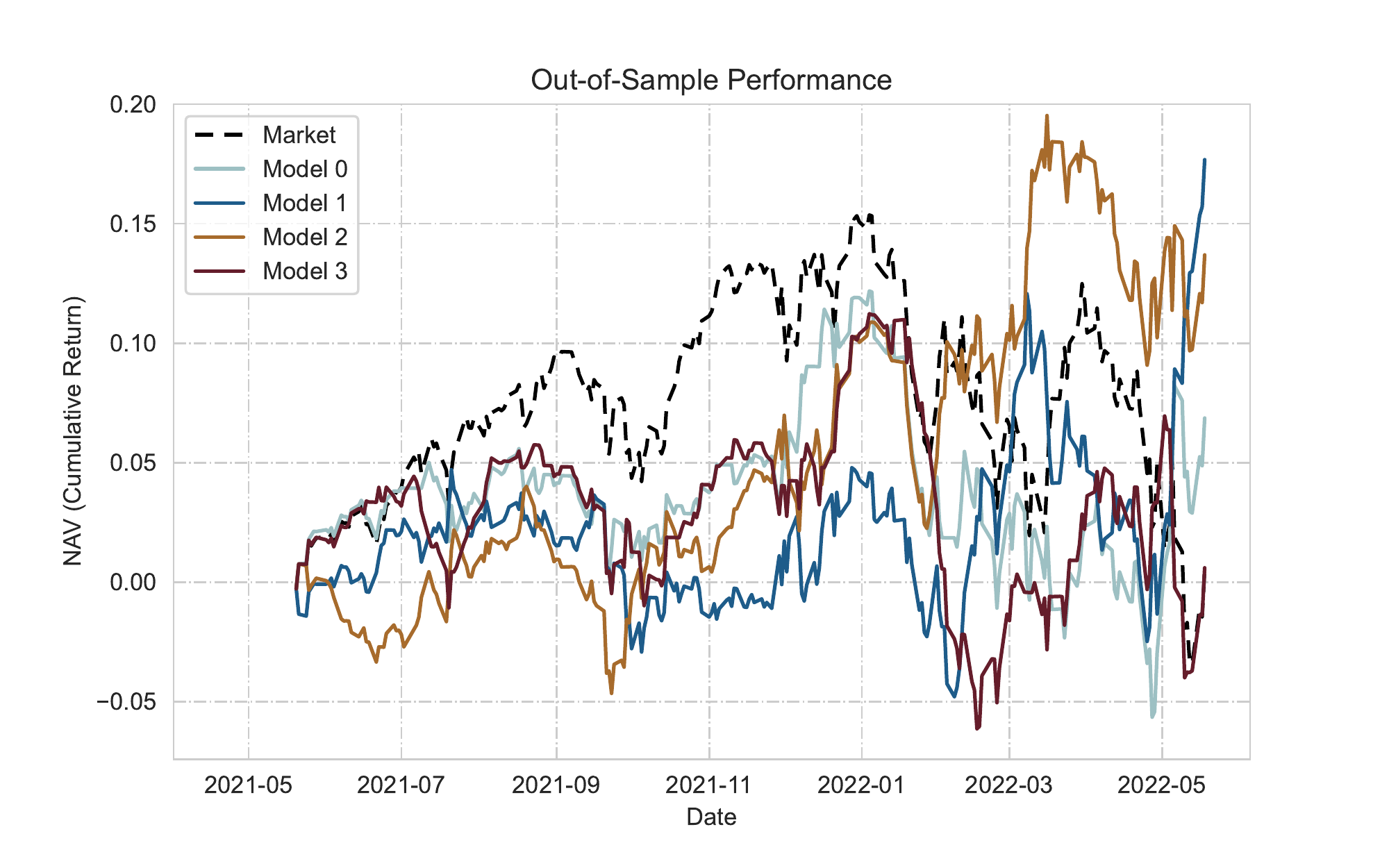}
     \end{subfigure}
    \caption{Performance of all models in-sample and out-of-sample {\bf with costs}.}
\end{figure}

\begin{table}
\scalebox{0.87}{
    \begin{minipage}{.49\linewidth}
        \renewcommand{\arraystretch}{1.5}
        \centering
        \caption{In-Sample Performance Keys}
        \begin{tabular}{| l | r | r | r |}
        \hline
                    & E(R)      & std(R)    & Sharpe    \\\hline
        Market      & 9.79\%    & 15.58\%   & 0.628     \\
        Model 0     & 15.19\%   & 19.77\%   & 0.768     \\
        Model 1     & 40.96\%   & 43.82\%   & 0.935     \\
        Model 2     & 29.54\%   & 36.54\%   & 0.809     \\
        Model 3     & 20.9\%    & 30.04\%   & 0.696     \\\hline
    \end{tabular}
    \end{minipage}
    }
\scalebox{0.87}{
    \begin{minipage}{.49\linewidth}
        \renewcommand{\arraystretch}{1.5}
        \centering
        \caption{Out-of-Sample Performance Keys}
        \begin{tabular}{| l | r | r | r | r | r |}
        \hline
                    & E(R)      & std(R)    & Sharpe    & MSE       & Accuracy  \\\hline
        Market      & 0.55\%    & 17.17\%   & 0.032     & 1.869     & 0.524     \\
        Model 0     & 6.87\%    & 16.57\%   & 0.414     & 1.810     & 0.452     \\
        Model 1     & 17.68\%   & 17.07\%   & 1.036     & 1.917     & 0.488     \\
        Model 2     & 13.69\%   & 16.82\%   & 0.814     & 1.937     & 0.468     \\
        Model 3     & 0.6\%     & 15.47\%   & 0.036     & 1.746     & 0.421     \\\hline
    \end{tabular}
    \end{minipage}
    }
\end{table}

\end{document}